\documentclass[11pt]{article}

\usepackage{acl}                       

\usepackage[english]{babel}
\babelfont{rm}{TeXGyreTermesX}         
\usepackage{latexsym}

\usepackage{amsmath,amssymb}
\usepackage{graphicx}
\usepackage{booktabs}
\usepackage{multirow}
\usepackage{xcolor}
\usepackage{colortbl}
\usepackage{enumitem}
\usepackage{microtype}
\usepackage{subcaption}
\usepackage{pifont}

\usepackage{listings}
\newcommand{\sys}{\textsc{MobileGym}}
\newcommand{\bench}{\textsc{MobileGym-Bench}}
\newcommand{\cmark}{{\color{green!60!black}\ding{51}}}
\newcommand{\xmark}{{\color{red!70!black}\ding{55}}}

\title{\sys{}: A Verifiable and Highly Parallel Simulation Platform for Mobile GUI Agent Research}

\author{
  \begin{tabular}{c}
    Dingbang Wu\textsuperscript{1,2,*} \quad Rui Hao\textsuperscript{3,1,*} \quad Haiyang Wang\textsuperscript{4} \quad Shuzhe Wu \quad Han Xiao\textsuperscript{5} \\
    Zhenghong Li\textsuperscript{1} \quad Bojiang Zhou\textsuperscript{1} \quad Zheng Ju\textsuperscript{1} \quad Zichen Liu\textsuperscript{1} \\
    Lue Fan\textsuperscript{1,5,6,8,$\dagger,\ddagger$} \quad Zhaoxiang Zhang\textsuperscript{1,7,8,9,$\dagger$}
  \end{tabular} \\
  \textsuperscript{1}Institute of Automation, Chinese Academy of Sciences (CASIA) \\
  \textsuperscript{2}School of Advanced Interdisciplinary Sciences, University of Chinese Academy of Sciences \\
  \textsuperscript{3}School of Artificial Intelligence, University of Chinese Academy of Sciences \\
  \textsuperscript{4}Peking University \quad
  \textsuperscript{5}The Chinese University of Hong Kong \quad
  \textsuperscript{6}CPII \\
  \textsuperscript{7}University of Chinese Academy of Sciences (UCAS) \\
  \textsuperscript{8}New Laboratory of Pattern Recognition (NLPR) \\
  \textsuperscript{9}State Key Laboratory of Multimodal Artificial Intelligence Systems (MAIS) \\
  \texttt{lue.fan@ia.ac.cn},\ \texttt{zhaoxiang.zhang@ia.ac.cn} \\[2pt]
  \textsuperscript{*}Equal contribution.\quad \textsuperscript{$\dagger$}Corresponding authors.\quad \textsuperscript{$\ddagger$}Project lead. \\
  Project page: \url{https://mobilegym.github.io}
}

\begin{document}
\setlength\titlebox{17.75\baselineskip}
\maketitle

\begin{abstract}
We present \sys{}, a \textbf{browser-hosted, lightweight, fully controllable simulation platform for everyday mobile use}, targeting \emph{interaction fidelity} without replicating proprietary backends. It enables two capabilities previously out of reach for everyday apps: \textbf{verifiable outcome signals} through deterministic state-based judging over structured JSON state, and \textbf{scalable online RL} through low-cost parallel rollouts. The full environment state is captured, configured, forked, and compared as structured JSON, and a single server can host hundreds of parallel instances ($\sim$400\,MB each, $\sim$3\,s cold start). A layered state model and a declarative task-definition framework keep state programmability and task creation practical at scale, and a single programmatic judging mechanism delivers both deterministic evaluation verdicts and dense RL rewards. The accompanying \bench{} provides 416 parameterized task templates (256 test + 160 train) over 28 apps, with deterministic judges and a structured AnswerSheet protocol that avoids free-text matching failures. In a Sim-to-Real case study, GRPO on Qwen3-VL-4B-Instruct gains $+$12.8\,pt on the 256-task test set, and on a 59-task real-device signal subset, real-device execution retains 95.1\% of the simulation-side training gain.
\end{abstract}

\section{Introduction}\label{sec:intro}

\begin{figure}[t]
  \vspace{7pt}
  \centering
  \includegraphics[width=\columnwidth]{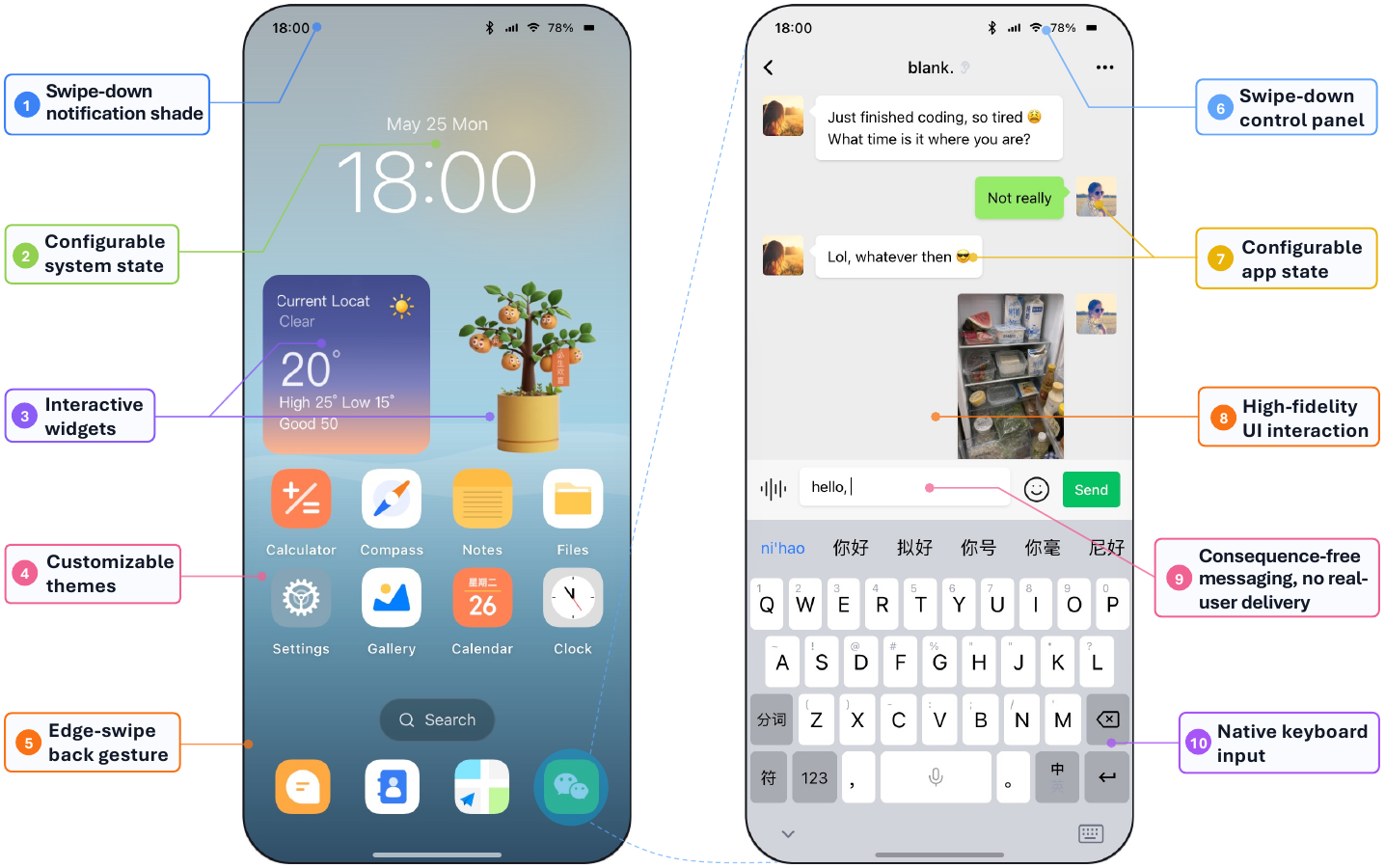}
  \caption{\textbf{Example screens from \sys{}.} Annotated launcher and messaging screens showing \sys{}'s configurable and sandboxed capabilities.}
  \label{fig:simulator-screenshots}
\end{figure}

\begin{figure*}[t]
  \centering
  \includegraphics[width=\textwidth]{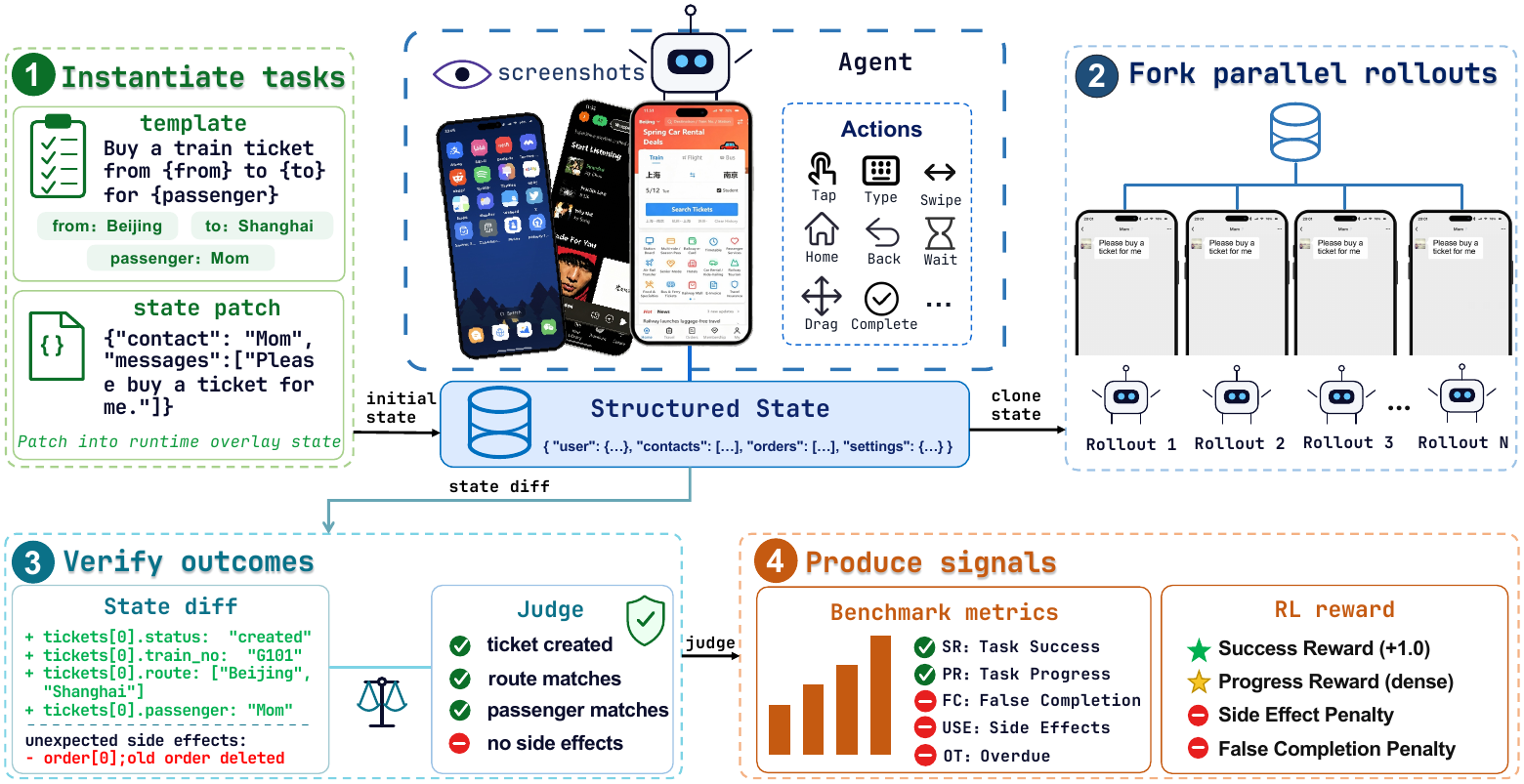}
  \caption{\textbf{End-to-end workflow of \sys{}.} A structured state supports task instantiation, parallel rollout forking, and state-diff verification. The resulting judgments are then converted into benchmark metrics and RL rewards.}
  \label{fig:teaser}
\end{figure*}

Mobile GUI agents have advanced rapidly in operating smartphones from screenshots and natural-language instructions~\cite{UITARS,AutoGLM,UIVenus15,GUIOw15,UIGenie}, yet current evaluation and training environments remain divided by a basic trade-off. Emulator-based environments such as AndroidWorld and AndroidLab~\cite{AndroidWorld,AndroidLab} offer repeatable evaluation but mainly cover system utilities and simple open-source apps, and scaling to online training requires many heavyweight emulator instances. Real-device benchmarks such as MobileBench-OL~\cite{MobileBenchOL} reach everyday apps, but live accounts, backend state, app-version drift, real-world consequences, and the cost of maintaining many devices and accounts make episodes difficult to control, reproduce, and parallelize. Neither route provides the combination needed for progress. First, environments need \textbf{verifiable outcome signals}, so benchmark verdicts and RL rewards are deterministic and grounded in actual task state rather than unreliable VLM judgments. Second, they need \textbf{scalable online training}: online RL has become an important capability driver for GUI agents~\cite{UIVenus15,UITARS2,AgentCPMGUI}, while offline trajectories struggle to cover dynamic GUI variations~\cite{MAIUI}.

The barriers are inherent to how everyday apps work. Everyday-app state is \textbf{unreadable}: internal state such as balances and orders is difficult to inspect through \texttt{adb} and accessibility trees, while VLM judges are intrinsically unreliable and further constrained by discrete screenshots that provide only partial evidence. It is \textbf{unwritable}: reproducible evaluation and online RL require resetting to known initial conditions, yet task-relevant state is split across proprietary storage, caches, and remote services, making desired states difficult to configure or restore. It is \textbf{unforkable}: large-scale online training benefits from parallel rollouts, and group-based methods such as GRPO further require multiple rollouts from identical initial states, yet live apps provide neither cheap replication nor state forking. Finally, many actions are \textbf{irreversible}, risking real messages, real transfers, or permanent account changes. These constraints make everyday apps structurally resistant to reproducible experimentation, even though they are natural targets for mobile-agent research. Scalability poses a further challenge: even for the apps emulators do support, each instance requires gigabytes of RAM, making large-scale parallel rollouts impractical on commodity hardware --- let alone for everyday apps that are resource-intensive or restrict emulator execution.

Yet GUI agents observe only screenshots and act through discrete actions, so a lightweight simulator with fully programmable state can be sufficient --- it only needs \emph{interaction fidelity}, producing realistic screens in response to agent actions. We introduce \sys{}, a browser-hosted Android-like simulation environment built on this principle. App data, OS state, and device context are represented as structured JSON, and the same mechanism makes state readable for deterministic outcome checking, writable for configuration and reset, forkable for parallel rollouts, and fully sandboxed for high-consequence actions. Agents observe only screenshots, while researchers retain full programmatic control. Each browser instance uses roughly 400\,MB of RAM and cold-starts in about 3\,s, enabling hundreds of parallel instances on a single server. For query tasks, a structured AnswerSheet protocol replaces brittle free-text matching with typed, GUI-submitted fields. Figure~\ref{fig:simulator-screenshots} shows example simulated screens, and Figure~\ref{fig:teaser} shows the end-to-end pipeline.

Our main contributions are:

\begin{itemize}
  \item \textbf{The \sys{} platform} (§\ref{sec:system}): a lightweight, browser-hosted Android-like simulation environment, including 12 everyday apps covering the major categories of daily mobile use and 16 system apps. Its modular app architecture and declarative task framework support easy extension, and a single machine can host hundreds of parallel instances.

  \item \textbf{Programmable state and verification mechanisms} (§\ref{sec:system:state}, §\ref{sec:bench:design}): full-environment state represented as structured JSON that supports deterministic judging, snapshot-based rollout forking, side-effect detection, and a typed AnswerSheet protocol that avoids free-text matching failures.

  \item \textbf{\bench{}} (§\ref{sec:bench}): 416 parameterized task templates (256 test + 160 train) covering major categories of everyday mobile use, with deterministic judges, empirically calibrated difficulty strata, and diagnostic metrics.

  \item \textbf{Empirical validation} (§\ref{sec:exp}): benchmark results across 9 agents (9.4\%--58.8\% SR), a GRPO training study gaining $+$12.8\,pt on the 256-task test set, a real-device study retaining 95.1\% of the simulated gain on a real-device signal subset, and a VLM-judge audit showing 10.2\% misjudgment.
\end{itemize}

\section{Related Work}\label{sec:related}

\begin{table*}[t]
  \centering\small
  \resizebox{\textwidth}{!}{%
  \begin{tabular}{lcccc>{\columncolor{gray!8}}c}
    \toprule
    & \textbf{AndroidWorld} & \textbf{AndroidLab} & \textbf{MobileWorld} & \textbf{MobileBench-OL} & \textbf{\sys{}} \\
    \midrule
    Runtime              & Emulator          & Emulator          & Emulator             & Real device        & Browser \\
    App types            & System + open-source & System + open-source & System + surrogates & Real everyday apps   & Simulated everyday + system \\
    Everyday app coverage   & \xmark            & \xmark            & Surrogates           & \cmark\ (real)     & \cmark\ (simulated) \\
    Apps / task units    & 20 / 116 templates & 9 / 138 instances & 20 / 201 tasks      & 80 / 1080 instances & 28 / 416 templates \\
    Verification         & adb programmatic  & UI-tree + LLM     & SQL + hooks          & XPath rules        & State-based programmatic \\
    Full environment state comparison & \xmark & \xmark & \xmark & \xmark & \cmark \\
    Snapshot \& restore  & App-data snapshot & AVD snapshot      & AVD snapshot         & \xmark             & JSON (ms-level) \\
    State customization  & Limited           & Limited           & Partial (SQL)        & \xmark             & Full (JSON) \\
    Online RL-ready      & Limited           & Limited           & Limited              & \xmark             & \cmark \\
    Memory per instance  & $\sim$4.5\,GB\textsuperscript{M} & $\sim$6\,GB\textsuperscript{R} & $\geq$4.5\,GB\textsuperscript{L} & N/A & $\sim$400\,MB\textsuperscript{M} \\
    Disk footprint       & $\sim$20\,GB\textsuperscript{M}  & $\sim$9\,GB\textsuperscript{R} & $\geq$20\,GB\textsuperscript{L}  & N/A & $\sim$50\,MB\textsuperscript{M} \\
    Cold start           & $\sim$78\,s\textsuperscript{M}   & ---                           & ---                            & N/A & $\sim$3\,s\textsuperscript{M} \\
    Parameterized tasks  & \cmark            & \xmark            & \xmark               & \xmark             & \cmark \\
    Sim-to-Real validation & N/A             & N/A               & N/A                  & N/A                & Validated \\
    \bottomrule
  \end{tabular}%
  }
  \caption{Comparison of mobile GUI agent benchmarks and infrastructures. Task-unit labels follow each benchmark's native counting unit. AndroidLab additionally releases 10.5k offline SFT trajectories, not counted here. \textit{Validated} denotes the real-device transfer study in §\ref{sec:exp:sim2real}, where 95.1\% of the simulation-side training gain on the 59-task signal subset is retained. Resource figures are \textsuperscript{M}measured by us, \textsuperscript{R}repository-reported configurations, or \textsuperscript{L}lower bounds inherited from a shared infrastructure baseline; measurement details are in Appendix~\ref{app:perf}.}
  \label{tab:comparison}
\end{table*}

\paragraph{Real-device and emulator route.} Existing mobile GUI agent environments run tasks on a heavyweight Android emulator or physical device and judge them externally, either through programmatic queries to interfaces such as \texttt{adb}, accessibility trees, UI-tree, or XPath rules, or through VLM-based screenshot judges. On system utilities and open-source apps, deterministic verification is feasible: AndroidWorld~\cite{AndroidWorld} judges 116 emulator tasks through \texttt{adb}, AndroidLab~\cite{AndroidLab} adds UI-tree matching with an LLM verifier for query-answer subtasks, and MobileWorld~\cite{MobileWorld} queries backend databases directly. A3~\cite{A3} targets 20 mainstream Google Play apps via Appium and adopts MLLM-as-judge to handle their dynamic content, trading determinism for coverage. MobileBench-OL~\cite{MobileBenchOL} runs 1080 tasks across 80 Chinese-language everyday apps on physical phones, the closest prior attempt at real everyday-app evaluation. Its XPath rules are brittle to unexpected popups and to minor app or backend updates, and the physical-device setup does not support parallel rollouts. All inherit the constraints discussed in §\ref{sec:intro}. Table~\ref{tab:comparison} compares representative environments.

\paragraph{Other mobile GUI benchmarks.} SPA-Bench~\cite{SPABench}, Mobile-Bench~\cite{MobileBench}, ProBench~\cite{ProBench}, MVISU-Bench~\cite{MVISUBench}, UI-NEXUS~\cite{UINEXUS}, and ColorBench~\cite{ColorBench} contribute task suites along axes orthogonal to environment infrastructure, and inform \bench{}'s taxonomy design (§\ref{sec:bench:taxonomy}).

\paragraph{Synthesis and trajectory-replay environments.} GUI-Genesis~\cite{GUI-GENESIS} reconstructs real apps as lightweight web environments from interaction trajectories with code-native rewards, but each environment covers only a single task trajectory. UISim~\cite{UISim} and ViMo~\cite{ViMo} adopt image-generation approaches. However, visual prediction errors can accumulate over long horizons, making these environments less suitable for RL with deterministic state transitions. OpenApps~\cite{OpenApps} focuses on reliability measurement with 6 FastHTML applications and shares the lightweight design philosophy of \sys{}, while pursuing a different goal.

\paragraph{Verifiable environments in other domains.} Beyond mobile, verifiable interactive environments have been built in the web domain (WebShop~\cite{WebShop}, WebArena~\cite{WebArena}, VisualWebArena~\cite{VisualWebArena}, WebGym~\cite{WebGym}, AutoWebWorld~\cite{AutoWebWorld}, InfiniteWeb~\cite{InfiniteWeb}), the desktop OS domain (OSWorld~\cite{OSWorld}, macOSWorld~\cite{macOSWorld}), and over simulated Python APIs (AppWorld~\cite{AppWorld}).

\paragraph{RL-based GUI agent training.} DigiRL~\cite{DigiRL} demonstrates a substantial advantage of online RL over SFT for device control. UI-TARS-2~\cite{UITARS2} deploys thousands of VMs to enable large-scale RL rollouts. UI-Venus-1.5~\cite{UIVenus15} introduces full-trajectory online RL with model fusion and achieves 77.6\% SOTA on AndroidWorld. GUI-Owl-1.5~\cite{GUIOw15} proposes the MRPO algorithm to address conflicts in multi-platform RL training. MobileGUI-RL~\cite{MobileGUIRL}, Mobile-R1~\cite{MobileR1}, UI-R1~\cite{UIR1}, GUI-R1~\cite{GUIR1} explore curriculum-style and R1-style training.

\section{The \sys{} Platform}\label{sec:system}

\begin{figure*}[t]
  \centering
  \includegraphics[trim=0 24bp 0 24bp,clip,width=\textwidth]{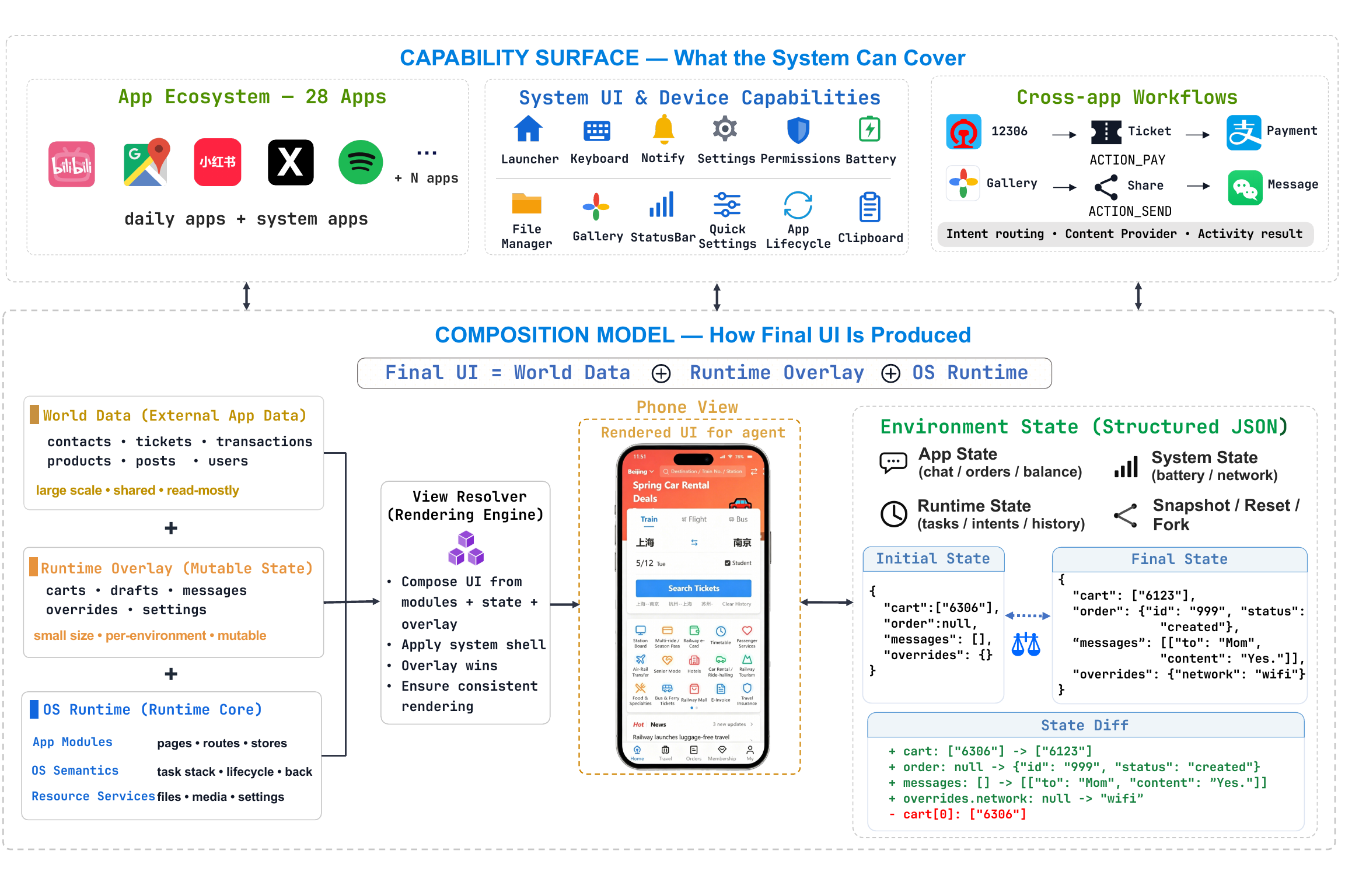}
  \caption{\textbf{System capabilities and state model of \sys{}.} App views are produced by composing read-mostly \emph{World Data}, a per-environment \emph{Runtime Overlay}, and the \emph{OS Runtime}. The resulting structured environment state supports snapshot/reset/fork and deterministic state-diff judging.}
  \label{fig:arch}
\end{figure*}

\sys{} is a browser-hosted Android-like simulation environment. Its app data, OS settings, and device properties are represented as explicit structured state, which the benchmark layer can configure, reset, snapshot, fork, and compare (Figure~\ref{fig:arch}).

\subsection{System Design}\label{sec:system:design}

\paragraph{Interaction fidelity target.} \sys{} does not aim to reproduce real everyday app backends or pixel-level Android internals. Its target is the interaction surface available to GUI agents: visual screens, touch and typing responses, navigation, cross-app handoffs, and task-relevant state transitions. As summarized in Figure~\ref{fig:arch}, this requires Android-like runtime mechanisms such as task stacks, keyboard, notification, and permission flows, shared resources, intent routing, content sharing, and back-key dispatch. These mechanisms are implemented in the browser over structured local state, making the same interaction semantics readable, writable, and forkable for evaluation and RL. Implementation details are in Appendix~\ref{app:system-detail}.

\paragraph{Layered state model.} The environment separates large, mostly read-only \emph{world data}, compact per-environment \emph{runtime state}, and OS runtime state. World data contains public entities such as posts and products, while runtime state contains data that can be changed by the agent, such as the current user's profile or app settings. Agent operations write only to runtime state, and views are produced by overlaying this layer on the read-only world data. Only runtime state is exposed for configuration, reset, judging, and comparison, keeping snapshots small and stable while preserving all agent-induced changes for full-environment state comparison.

\paragraph{Declarative navigation specification.} The UI navigation of every app is modeled as a declarative finite-state machine, built at development time into a per-app specification file. The same file drives runtime navigation and static analysis, including task-trajectory enumeration, and auto-generation of new tasks. The formal definition and guard syntax are provided in Appendix~\ref{app:efsm}.

\paragraph{Interface and extensibility.} The Benchmark layer maps agent outputs to a unified 17-action abstraction (Appendix~\ref{app:action-space}), executes actions through Playwright with coordinates normalized to $[0,1000]$, and returns only screenshots. On the app side, \sys{} provides a repeatable module architecture that separates UI pages, app-local runtime state, declarative navigation, replaceable default data, and world data, allowing new apps and features to reuse the shared OS lifecycle, reset, snapshot, rollout, and judging interfaces (Appendix~\ref{app:system-detail}).

\subsection{State Programmability}\label{sec:system:state}

\paragraph{Verifiable outcome signals.} Task success is judged by \textbf{programmatic state verification}: each task has a deterministic judge that inspects environment state. This provides deterministic, fine-grained outcome signals without unreliable VLM judgments.

\paragraph{State serialization and multi-instance replication.} The full environment state can be serialized as structured JSON and restored on demand, enabling exact reset and forking from any snapshot, supporting RL methods such as GRPO. For irreversible operations (transfers, deactivation, deletions, etc.), the consequence-free simulator allows full restoration after each trajectory.

\paragraph{Full environment state comparison.} The fully structured state enables full-environment state comparison between an episode's initial and terminal states, reporting any mutation outside the task's expected outcome as an \emph{unexpected side effect}. For personal mobile agents, this distinction is critical: an agent may complete the requested goal while, for example, sending an unintended message. This mechanism defines the \textbf{Unexpected Side Effects} metric (§\ref{sec:bench:eval}). Existing programmatic mobile benchmarks do not provide this environment-wide signal, and VLM judges can only approximate it from screenshots without deterministic guarantees.

\section{The \bench{}}\label{sec:bench}

\bench{} is a suite of 416 parameterized task templates (256 test + 160 train, strictly disjoint) built on top of the \sys{} platform. It covers major categories of everyday mobile use.  Detailed information about the 28 apps and representative task examples is listed in Appendix~\ref{app:apps}.

\subsection{Task Taxonomy}\label{sec:bench:taxonomy}

Prior task taxonomies often couple unrelated dimensions, such as mixing app count with subtask count~\cite{MobileBench}. We factor the task space along four orthogonal axes:
\begin{itemize}
  \item \textbf{Scope} --- how many apps a task involves. S1: single-app, S2: two-app, S3: three or more.
  \item \textbf{Objective} --- what the task asks for. Operate: state-changing actions, query: information retrieval, hybrid: both.
  \item \textbf{Composition} --- how subtasks are structured. Atomic: a single action, sequential: an ordered chain, transfer: cross-app handoff, deep-dive: multi-step drill-down.
  \item \textbf{Difficulty} --- how hard the task is for current models. L1: easy, L2: moderate, L3: hard, L4: very hard. Calibrated post-hoc using eight reference models, details in §\ref{sec:bench:difficulty}.
\end{itemize}
Each task is additionally annotated with 1--4 capability tags from a 13-tag vocabulary. The full taxonomy and tag definitions are provided in Appendix~\ref{app:taxonomy}.

\subsection{Task Design}\label{sec:bench:design}

Two design choices shape the task suite: parameterized instantiation for diversity, and AnswerSheet fields for query-task judging reliability.

\paragraph{Parameterized task instantiation.} The 416 entries in \bench{} are \textbf{templates}, not fixed instances. Each template is instantiated at runtime through three sources of variation: \textbf{(i) instruction variation}, where semantically equivalent goal phrasings are sampled; \textbf{(ii) parameter sampling}, where slot values are drawn from curated sets, numeric ranges, or the current environment state; and \textbf{(iii) environment configuration}, where app state such as contacts or order history is set through shared base data or per-task injections before rollout. Together, these variations reduce memorization of fixed instances and expand task diversity without requiring each instance to be authored separately. Across finite parameter ranges, they yield over 27,000 distinct task instances, not counting templates with continuous ranges that contribute unbounded additional instances.

\begin{figure}[t]
  \centering
  \includegraphics[width=\columnwidth]{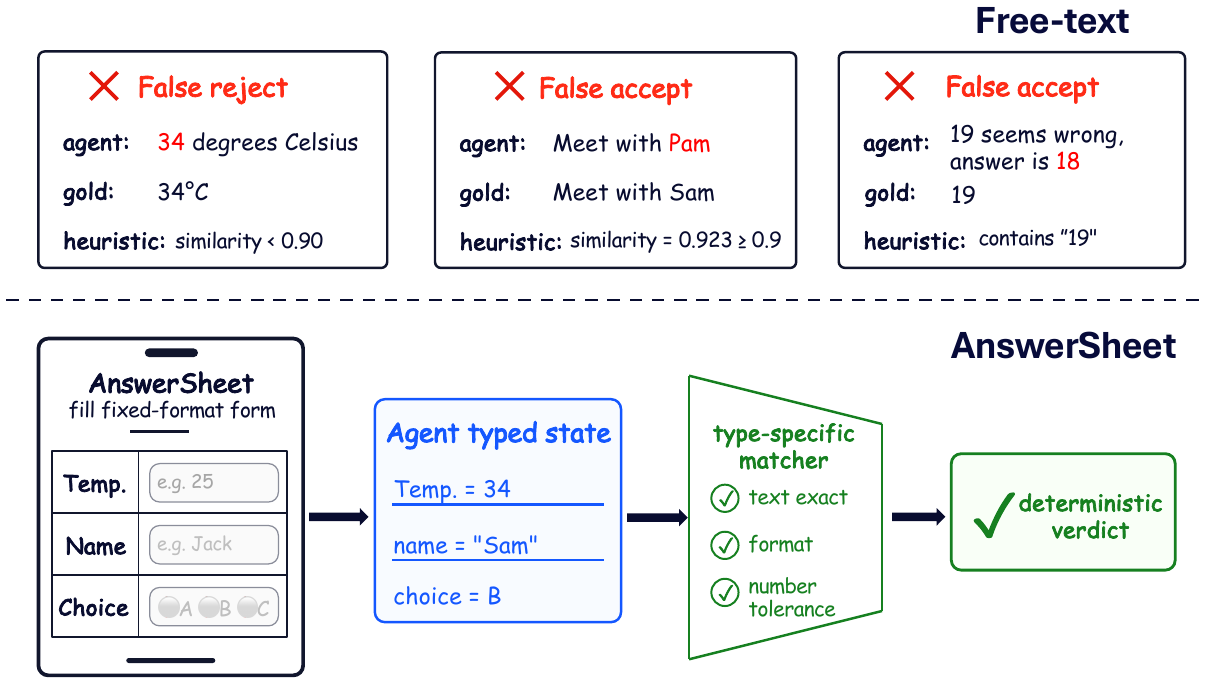}
\caption{\textbf{AnswerSheet protocol.} Free-text heuristics can reject equivalent answers or accept leaked reasoning that contains the gold answer. AnswerSheet instead uses GUI form filling and type-specific checks over typed fields.}
  \label{fig:answersheet}
\end{figure}

\paragraph{The AnswerSheet protocol.} Existing mobile benchmarks often judge free-text query answers with string-similarity or substring heuristics~\cite{AndroidWorld,MobileWorld}, which can reject equivalent phrasings or falsely accept answers that leak reasoning text containing the gold answer. \sys{} instead moves answer submission into the GUI: query tasks end with the agent filling an \textbf{AnswerSheet} form whose fields declare types and show format hints (Figure~\ref{fig:answersheet}). This preserves a natural form-filling interaction for GUI-specialized agents, while the submitted typed state is checked by type-specific matchers such as exact text, numeric tolerance, format, or choice checks. Details are in Appendix~\ref{app:answersheet}.

\subsection{Evaluation Protocol}\label{sec:bench:eval}

We report success, progress, termination, and side-effect diagnostics under fixed step budgets.

\paragraph{Metrics.} \textbf{Success Rate (SR)}, the fraction of tasks judged successful, is the primary metric. Diagnostics include \textbf{Progress Rate (PR)}, the fraction of subtasks passed; \textbf{False Complete (FC)}, episodes where the agent declares completion without success; \textbf{Unexpected Side Effects (USE)}, episodes with unexpected state changes; and \textbf{Overdue Termination (OT)}, episodes where the agent reaches the goal but continues until truncation.

\paragraph{Execution setup.} The simulator is reset before each task, and agents observe only screenshots. Each task is assigned one of four step budgets (15, 30, 45, or 60), manually verified to comfortably exceed its optimal completion length. Tasks with AnswerSheet receive an additional 15-step budget.

\subsection{Model-Calibrated Difficulty Strata}\label{sec:bench:difficulty}

Motivated by benchmark-curation precedents such as BBH, which identifies hard tasks using prior model and human-rater performance~\cite{BBH}, four difficulty levels are assigned by \textbf{post-hoc empirical calibration}. We evaluate eight reference models\footnote{Gemini 3.1 Pro, Doubao-Seed-2.0-Pro, Qwen3.6-Plus, AutoGLM-Phone-9B, UI-TARS-1.5-7B, UI-Venus-1.5-8B, GUI-Owl-1.5-8B-Think, Step-GUI-4B.} on the test set and stratify tasks by mean SR and PR: L1 (SR${\geq}$75\%, PR${\geq}$75\%, $n{=}20$), L2 (remaining tasks with SR${\geq}$25\%, PR${\geq}$50\%, $n{=}73$), L3 (remaining tasks with SR${>}$0, PR${\geq}$25\%, $n{=}83$), and L4 (otherwise, $n{=}80$). These are diagnostic strata rather than intrinsic labels, and the calibration excludes Qwen3-VL-4B-Instruct and its fine-tuned variants used in §\ref{sec:exp:sim2real}. A reference-model robustness check is reported in Appendix~\ref{app:panel-sensitivity}.

\section{Experiments}\label{sec:exp}

We evaluate 9 agents on \bench{} (Table~\ref{tab:main-results}). Open-source models use 4 trials with re-sampled parameters; proprietary models use one due to API cost, with one additional run for Gemini 3.1 Pro, the strongest model, to estimate variation.

\subsection{Benchmark Results}\label{sec:exp:bench}

\begin{table*}[t]
  \centering\small
  \setlength{\tabcolsep}{4pt}
  \begin{tabular*}{\textwidth}{@{\extracolsep{\fill}}lcc|cccc|ccc}
    \toprule
    & \multicolumn{2}{c|}{\textbf{Overall (\%)}} & \multicolumn{4}{c|}{\textbf{Difficulty SR (\%)}} & \multicolumn{3}{c}{\textbf{Diagnostics (\%)}} \\
    \textbf{Model} & \textbf{SR} & \textbf{PR} & \textbf{L1 (20)} & \textbf{L2 (73)} & \textbf{L3 (83)} & \textbf{L4 (80)} & \textbf{FC} & \textbf{OT} & \textbf{USE} \\
    \midrule
    \rowcolor{gray!8} \multicolumn{10}{l}{\textit{Proprietary models}} \\
    Gemini 3.1 Pro       & 58.8 \scriptsize{$\pm$1.4} & 72.1 &  97.5 & 83.6 & 63.3 & 21.9 & 34.0 & 0.2 &  5.5 \\
    Doubao-Seed-2.0-Pro  & 52.0$^\dagger$             & 63.6 & 100.0 & 93.2 & 48.2 &  6.2 & 33.6 & 0.4 &  4.7 \\
    Qwen3.6-Plus         & 45.7$^\dagger$             & 59.2 & 100.0 & 78.1 & 44.6 &  3.8 & 34.0 & 0.4 & 14.5 \\
    \midrule
    \rowcolor{gray!8} \multicolumn{10}{l}{\textit{Open-source GUI-specialized models}} \\
    AutoGLM-Phone-9B     & 20.0 \scriptsize{$\pm$1.3} & 35.3 & 86.2 & 33.6 &  9.6 & 1.9 & 39.6 & 0.6 & 12.6 \\
    UI-TARS-1.5-7B       & 13.8 \scriptsize{$\pm$1.7} & 26.3 & 77.5 & 21.9 &  3.0 & 1.6 & 38.6 & 0.2 & 11.0 \\
    UI-Venus-1.5-8B      & 15.4 \scriptsize{$\pm$2.4} & 28.3 & 85.0 & 21.9 &  6.0 & 1.9 & 22.9 & 0.5 &  7.7 \\
    GUI-Owl-1.5-8B-Think & 15.1 \scriptsize{$\pm$0.9} & 28.8 & 76.2 & 26.0 &  4.2 & 1.2 & 30.4 & 0.9 & 14.1 \\
    Step-GUI-4B          & 12.9 \scriptsize{$\pm$1.1} & 25.7 & 83.8 & 17.8 &  2.4 & 1.6 & 37.0 & 0.8 &  7.6 \\
    \midrule
    \rowcolor{gray!8} \multicolumn{10}{l}{\textit{Open-source generalist models}} \\
    Qwen3-VL-4B-Instruct          &  9.4 \scriptsize{$\pm$0.6} & 20.1 & 71.2 & 12.3 &  0.6 & 0.3 & 15.9 & 0.4 & 10.0 \\
    \bottomrule
  \end{tabular*}
\caption{Main results on the \bench{} test set (256 tasks). Overall reports Success Rate (SR) and Progress Rate (PR); Difficulty SR reports SR within calibrated difficulty strata L1--L4, with task counts in parentheses; Diagnostics report False Complete (FC), Overdue Termination (OT), and Unexpected Side Effects (USE). $\pm$ denotes standard deviation across trials; $^\dagger$ marks single-run results. See §\ref{sec:bench:eval} for metrics and Appendix~\ref{app:results-detail} for details.}
  \label{tab:main-results}
\end{table*}

Two observations stand out from Table~\ref{tab:main-results}. Additional experimental results are in Appendix~\ref{app:results-detail}.

\paragraph{Difficulty stratification.} SR decreases monotonically from L1 to L4 for all 9 models, while overall SR spans 9.4\%--58.8\%, giving a 6$\times$ performance range without top saturation or bottom floor effects. L1 already separates proprietary and open-source agents, and L4 acts as the frontier discriminator: only Gemini 3.1 Pro retains meaningful performance at 21.9\%, while all other proprietary models reach at most 6.2\% and all open-source GUI specialists at most 1.9\%.

\paragraph{Unexpected side effects.} USE captures unintended agent operations that modify state unrelated to the task. It does not simply decrease with model capability: across the 9 models it ranges from 4.7\% to 14.5\%, and even open-source GUI specialists with similar SRs (12.9--15.4\%) differ nearly 2$\times$ in USE (7.6--14.1\%). This diagnostic is enabled by \sys{}'s full-environment state comparison. Screenshot or UI-tree judges cannot reliably expose off-target changes hidden in app-internal or backend state.

\subsection{Sim-to-Real Transfer}\label{sec:exp:sim2real}

We view this real-device experiment as an existence proof that
training in \sys{} can produce behavior that survives real-device
execution, not as a comprehensive sim-to-real study.

We fine-tune Qwen3-VL-4B-Instruct with GRPO~\cite{GRPO} on \sys{}'s 160-task train set for 10 steps, using a single node with 3 RTX Pro 6000s and 96 parallel environment instances. Key hyperparameters are $\mathrm{lr}=10^{-6}$, group size $k{=}8$, batch size $bs{=}12$, KL 0.01, DAPO~\cite{DAPO}-style asymmetric clip-higher (0.2/0.28). The reward is a PR-shaped dense signal, with multiplicative penalties for AnswerSheet error, side effects, false completion, and overdue/post-success abort. Details are provided in Appendix~\ref{app:exp-config}.

\paragraph{Training gains on the simulation side.} Training raises overall SR from 9.4\% to 22.2\% (\textbf{$+$12.8\,pt}) on the 256-task \bench{} test set. Broken down by difficulty, SR changes from 71.2\% to 92.5\% on L1, 12.3\% to 37.7\% on L2, 0.6\% to 11.7\% on L3, and 0.3\% to 1.2\% on L4. The lift is largest on L2 and nearly flat on L4, suggesting that training is most effective where the base model already exhibits moderate capability, while the hardest tasks remain capacity-limited. The trained 4B model surpasses the 9B AutoGLM-Phone-9B on L1--L3, while both remain near zero on L4.

\paragraph{Real-device evaluation design.} We evaluate on a Redmi Note 12 Turbo ($1080\times2400$). We stratify the 256-task test set by the base/trained models' pass counts over four simulator rollouts: Uplift (base $\leq$1, trained $\geq$3; 26 tasks), Stable-pass (both $\geq$3; 21 tasks), Mid (all remaining cases; 20 tasks), Regression (base $\geq$3, trained $\leq$1; 0 tasks), and Stable-fail (both $\leq$1; 189 tasks). The three signal buckets (Uplift, Mid, Stable-pass) contain 67 tasks, of which 59 can be safely and equivalently run on the real device after excluding 8 tasks involving unreproducible account state or irreversible operations. Running all 189 Stable-fail tasks on a single real device serially would be costly  and manual state restoration, while these tasks, by definition, exhibit no simulation-side training gain to transfer. We therefore randomly sample 15 as a negative-control check. The real-device setting differs from simulation in UI details, app data, real-app variability, and task entities such as contacts or POIs. Details in Appendix~\ref{app:sim2real-buckets}.

\begin{figure}[t]
  \centering
  \includegraphics[width=\columnwidth]{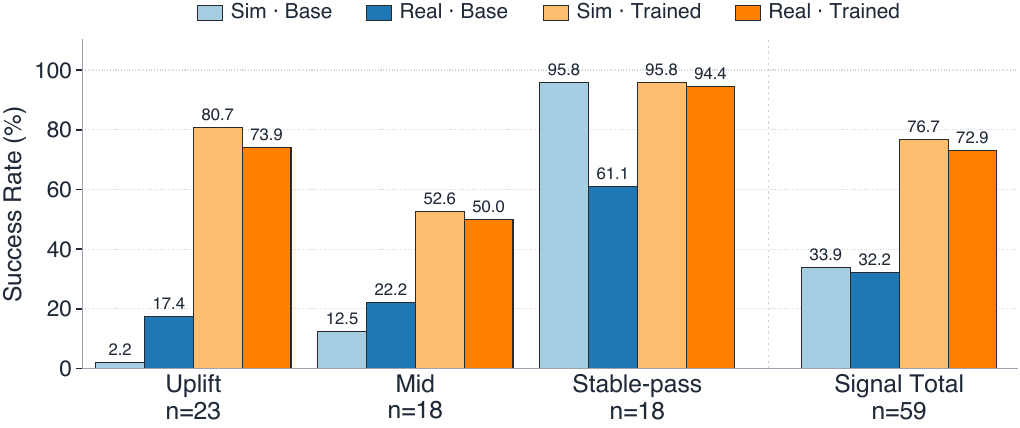}
  \caption{\textbf{Sim-to-Real transfer of GRPO training gains.} Per-bucket Success Rate on the 59-task signal-bucket subset and the overall Signal Total. In the legend, Sim/Real denotes the evaluation environment and Base/Trained denotes before/after GRPO. Sim columns are 4-seed averages, Real columns are pass@1 and all manually audited (Appendix~\ref{app:vlm-audit}).}
  \label{fig:sim2real}
\end{figure}

\paragraph{Results.} On the 59-task signal-bucket subset, training raises the real-device pass rate from 32.2\% to 72.9\% (\textbf{$+$40.7\,pt}), closely matching the simulation-side increase from 33.9\% to 76.7\% (\textbf{$+$42.8\,pt}). This corresponds to \textbf{95.1\% retained gain}. The absolute sim--real gaps are small for both the base model (1.7\,pt) and the trained model (3.8\,pt) (Figure~\ref{fig:sim2real}). The 15 randomly sampled Stable-fail tasks yield 0/15 success for both models on the real device, consistent with simulation. Because the real-device environment and task entities differ from the simulator, the lift more plausibly reflects transferable policies than memorization.

\paragraph{Trajectory-length check.} Successful trajectories on \texttt{operate} tasks have similar lengths in simulation and on device: 5.00 vs.\ 6.03 steps for the base model and 10.08 vs.\ 12.20 for the trained model. We exclude \texttt{query}/\texttt{hybrid} tasks because AnswerSheet adds simulator-only steps. Details in Appendix~\ref{app:fidelity-breakdown}.

\paragraph{Failure-recovery example.} In a real-device Reddit post-creation task, the selected community required a flair before the Post button could be enabled. The base model looped on the disabled button until the step budget was exhausted, while the trained model identified the missing flair requirement and succeeded in 22 steps (Appendix~\ref{app:case-study}).

\paragraph{VLM judge error analysis.} Manual review of all 118 signal-bucket real-device trajectories identifies 12 errors for Qwen3.6-Plus (5/59 base, 7/59 trained; 10.2\% overall). Re-judging the same trajectories with GPT-5.4~\cite{GPT54} also yields 12/118 errors, although on a partially different subset. These results suggest that VLM-judge errors are not specific to a single judge model, while programmatic state verification avoids this failure mode. Detailed counts are in Appendix~\ref{app:vlm-audit}.

\subsection{Efficiency Analysis}\label{sec:exp:efficiency}

\sys{} runs in the browser with roughly 1/10 the memory and under 1/100 the disk footprint of emulator-based setups (Table~\ref{tab:comparison}). In our measurements, 256 parallel instances on one server used $<$10\% CPU and $\sim$100\,GB RAM, completing a full 256-task benchmark evaluation in about 6 minutes. By contrast, MAI-UI~\cite{MAIUI} reports requiring 10 bare-metal cloud servers (960 vCPUs, 3{,}840\,GB RAM total) to reach 512 parallel Android-emulator instances for online RL. This single-node parallel capacity makes concurrent online RL (§\ref{sec:exp:sim2real}) feasible without dedicated cluster infrastructure. Appendix~\ref{app:cost} quantifies the API cost of using VLM judges instead of code-level judging.

\section{Conclusion}\label{sec:conclusion}

\sys{} turns everyday mobile use into a fully controllable simulation platform for GUI agent research. By targeting interaction fidelity rather than replicating proprietary backends, \sys{} makes everyday-app state readable for deterministic verification, writable for reset and configuration, forkable for parallel online RL, and consequence-free for high-risk operations. The \bench{} suite operationalizes this environment with 416 parameterized task templates, calibrated difficulty strata, structured AnswerSheet-based evaluation, and diagnostic metrics including unexpected side effects. Experiments across 9 agents show substantial headroom on everyday mobile tasks, and the Sim-to-Real study shows that most of the simulation-side training gain transfers to real-device execution. The same controllable infrastructure could also be used for safety-alignment research, robustness testing, and training-data generation (Appendix~\ref{app:broader-uses}). More broadly, \sys{} shows that interaction-fidelity simulation can make everyday mobile tasks available for reproducible research and scalable training, without relying on real accounts, device farms, or proprietary backends.

\section*{Limitations}

\paragraph{Visual appearance modeling.} Observed visual differences between \sys{} and the corresponding real apps mainly include subtle layout details, animations, and certain app-specific icons. The Sim-to-Real experiment (§\ref{sec:exp:sim2real}) provides one quantitative datapoint that this level of visual similarity can support behavioral policy transfer. Tasks that depend heavily on recognizing exact app-specific icons may still be affected during transfer.

\paragraph{Backend and dynamic-content modeling.} \sys{} models agent-facing interaction semantics rather than real service backends. Server-driven content such as ads, pop-ups, recommendation feeds, and real-time messages is represented as controllable JSON state, which favors deterministic reset, reproducible evaluation, and stable RL reward signals. This design does not capture backend-only or stochastic phenomena such as live recommendation dynamics, fraud checks, latency spikes, or server-side policy changes unless they are explicitly modeled as controllable state. Controllable dynamic-content injection is architecturally supported and left for future study.

\paragraph{Functional coverage of simulated apps.} Each simulated app implements the main everyday-use scenarios of its real counterpart rather than the full feature surface. Less common features remain out of scope. Expanding within-app coverage is future work.

\section*{Ethical Considerations}

\sys{} is a \textbf{fully sandboxed} research infrastructure. All simulation of commercial apps is disconnected from any real service, real account, real funds, or personal data.

\paragraph{Legality of commercial-app simulation.} The commercial apps reproduced in \sys{} are used only for academic research and model evaluation. Their trademarks, brand names, and visual elements remain the property of their respective owners. \sys{} does not reuse or distribute any official code or client components. The simulator UI is independently implemented with LLM-assisted programming and, due to the limits of model-based reproduction, differs from the real apps in pixel-level visual details. The environment runs in the browser, is offline, and never touches real accounts or funds. We do not claim any commercial or derivative use.

\paragraph{Double-edged nature of evaluating high-risk operation capabilities.} The high-risk subset (Appendix Table~\ref{tab:high-risk-results}) consists of 14 tasks: 7 standalone payment operations (Payment) and 7 high-consequence tasks drawn from Test256 (Test256-Risk, including account deactivation, large transfers, bulk deletions, etc.). Gemini 3.1 Pro reaches 64.3\% on Payment and 71.4\% on Test256-Risk, while smaller open-source GUI specialists remain at $\leq$10.7\% on Payment. Trajectory inspection finds no evidence of explicit refusal in either tier: frontier models attempt the operation and largely succeed, whereas open-source models attempt but fail. \textbf{We explicitly state that this is a report of execution capability, not an endorsement of such autonomous operations.} Under the current training paradigm, ``execution capability'' and ``operational caution'' are not decoupled. Frontier models, when instructed, execute irreversible operations with high success and no intrinsic caution gating. We argue that capability evaluation must be paired with safety alignment. The ability of \sys{} to simulate irreversible operations provides a no-real-risk testing infrastructure for follow-up safety-alignment research, which is an important part of its value.

\paragraph{Misuse risk and mitigation.} Any GUI-agent training infrastructure could potentially be used to automate malicious behavior. \sys{} is, by design, a research tool for capability evaluation and safety research, not a production deployment. We encourage using \sys{} for defensive research as well—safety alignment, prompt-injection robustness, and refusal training.

\paragraph{Societal impact.} The safe-simulation properties of \sys{}—zero-consequence operations, one-click reset, built-in difficulty levels—naturally make it suitable for \textbf{digital-literacy education}. Learners can repeatedly practice tasks such as contact lookup, mobile payment, and ticket booking in a fully simulated environment without any real consequences. We encourage the community to explore positive social applications of \sys{} in digital inclusion, customer-service training, and AI-safety education.

\section*{Acknowledgments}

This work was supported by the Jiangsu Science and Technology Major Project (BG2025035) and in part by the National Natural Science Foundation of China (Nos. 62320106010, 62506356, and 62536010). This work was also supported by the Beijing Natural Science Foundation (Nos. L257004 and L257015).

We thank the reviewers and area chairs for their constructive feedback. Cursor, Claude Code, and Codex supported code implementation.

\bibliography{references}


\appendix

\section{System Implementation Details}\label{app:system-detail}

This appendix provides implementation details that are omitted from §\ref{sec:system} for space.

\subsection{TaskManager and the Activity stack}

The application life-cycle management of \sys{} mirrors the \texttt{ActivityTaskManager} of Android. Each app runs in its own \textbf{Task}, and each Task maintains its own \textbf{Activity stack}. The TaskManager handles requests in a Reducer pattern: \texttt{LAUNCH\_APP} (start a new Task or reuse an existing one), \texttt{GO\_HOME} (return to desktop), \texttt{SHOW\_RECENTS} (open the recent-tasks list), \texttt{CLOSE\_TASK} (close and destroy React components), \texttt{PUSH\_ACTIVITY}/\texttt{POP\_ACTIVITY} (Activity push / pop).

App keep-alive is implemented by setting backgrounded Activity containers to \texttt{display:none} rather than unmounting the React component. The React state tree is therefore preserved, so when the user switches back, the interface can be restored without rebuilding the component tree. For example, when a user types half a draft in WeChat, switches to Alipay to complete a transfer, and then switches back to WeChat, the draft remains available. This behavior is important for the interaction fidelity of cross-app tasks. The TaskManager uses a non-persistent store, so a browser refresh is equivalent to a device reboot that returns to the desktop.

\subsection{State layer aligned with the data model of Android}

\begin{table}[!htbp]
  \centering\small
  \setlength{\tabcolsep}{3pt}
  \resizebox{\columnwidth}{!}{%
  \begin{tabular}{llc}
    \toprule
    \textbf{Android tier} & \textbf{\sys{} implementation} & \textbf{Persisted} \\
    \midrule
    Build Props        & OsStateStore.build  & \checkmark \\
    Settings           & OsStateStore.settings  & \checkmark \\
    Hardware/Sensor    & + Managers   & \checkmark \\
    App Data           & createAppStore (per-app)  & \checkmark \\
    \midrule
    \textit{Runtime}    & Volatile stores  & $\times$ \\
    \bottomrule
  \end{tabular}%
  }
  \caption{Android data model $\to$ \sys{} state-layer mapping}
  \label{tab:android-layers}
\end{table}

The persistence policy follows the core semantics ``browser refresh = device reboot'': user data is preserved across refreshes, while runtime state is reset on refresh. All stores are created through a unified factory and automatically registered in a global registry, so the entire environment state can be snapshotted or reset with a single call. This infrastructure supports bit-level consistent reset and programmatic state verification.

Hardware-state constraint logic is encapsulated in a Manager layer: \texttt{ConnectivityManager} (handles airplane-mode $\to$ cascading WiFi/Bluetooth/cellular shutdown), \texttt{BatteryManager} (battery and charging), \texttt{AudioManager} (volume and DND), and \texttt{DisplayManager} (brightness and zoom). A Manager acts as a write-side facade for the OsStateStore: when the Benchmark layer injects \texttt{airplane\_mode:true} through \texttt{setState()}, the Manager automatically cascades the dependent state.

\subsection{Cross-app communication}

\textbf{Intent system.} Each app declares the Intent types it can handle in its manifest. When the \texttt{IntentResolver} receives an Intent, it scans every manifest for matches. With a unique match, it transitions directly; with multiple matches, it displays a Chooser. Cross-app calls with callbacks (\texttt{startActivityForResult}) are also supported.

\textbf{ContentProvider.} Shared data is accessed via the \texttt{content://} protocol; we currently implement Contacts, Sms, and Media providers, supporting CRUD operations and change notifications.

\textbf{BroadcastBus.} System-level events are dispatched through a broadcast mechanism that is semantically aligned with \texttt{sendBroadcast} / \texttt{registerReceiver} in Android.

\subsection{Back-key dispatch}

The BackDispatcher implements priority-chain dispatch: permission dialog (1000) $>$ system shade (800) $>$ keyboard (700) $>$ app page (100) $>$ return-to-desktop (0). Back-key events are propagated from the highest priority downwards, and the first handler that returns \texttt{true} consumes the event. A frame-level deduplication mechanism (back lock) prevents the edge gesture and a backdrop click from double-triggering within the same frame.

\subsection{Standardized App-layer architecture}

Every app follows the same directory structure: \texttt{manifest.ts} (declares the app identity), \texttt{*App.tsx} (entry component, using MemoryRouter), \texttt{state.ts} (Zustand-based store), \texttt{navigation.declaration.ts} (declarative navigation specification), and \texttt{data/defaults.json} (replaceable initial data). At compile time, the OS uses Vite \texttt{import.meta.glob} to scan \texttt{apps/*/manifest.ts} and \texttt{system/*/manifest.ts}. This design supports \textbf{zero-registration auto-discovery}: once an app module provides a manifest, the OS automatically places it on the desktop and matches its Intents, removing OS-side registration work; implementing the app itself still requires page components, state stores, navigation declarations, and data.

\subsection{Input injection and coordinate transformation}

The screenshot pixel coordinates observed by the agent go through the following transformation chain: (1) screenshot pixel $\to$ CSS viewport coordinates, scaled by the ratio between screenshot resolution and viewport size; (2) the target DOM element is located through \texttt{document.elementFromPoint}; and (3) standard PointerEvent / TouchEvent sequences are generated and dispatched into the React event system. For apps that declare a \texttt{designViewportWidth} (e.g.\ the 412\,px design width of WeChat), an additional inverse transform is needed when CSS zoom is in effect.

\subsection{End-to-end App, Task, and Judge development workflow}

\paragraph{App development and validation.}
Development begins with human exploration of the corresponding real App. Developers examine its functional structure, state-dependent entry points, action branches, navigation and back behavior, and user-observable outcomes, and organize these observations into implementation requirements. Coding Agents then implement pages, structured states, business logic, synthetic data, resources, and a declarative navigation specification under \sys{}'s shared architecture and platform contracts. The standardized module structure, automatic discovery, reusable system services, and Vite hot-module replacement (HMR, where code edits become effective in $<$1 second) reduce repeated infrastructure work.

Automated checks verify consistency between navigation declarations and their implementations and generate navigation graphs and reachable paths. Human developers subsequently operate the simulated App and compare its major implemented functionality against observations from the real App, including functional logic, state changes, navigation, back behavior, and cross-App outcomes. Discrepancies are reported to Coding Agents for iterative correction. Together, the automated checks and human comparative testing verify the behavioral agreement of the implemented, user-observable functionality.

\paragraph{Task and Judge construction.}
Task construction has three stages. First, human developers derive objectives from realistic mobile scenarios and specify the natural-language instruction, parameters, initial state, and expected outcome. Second, Coding Agents inspect the relevant structured-state interfaces and translate the task intent into instruction templates, task configurations, success and progress conditions, and permitted state changes. Simple state-changing tasks reuse shared base classes and declare a small number of target states; more complex tasks compose multiple conditions or introduce custom Judge logic. Automated tests cover successful execution, failure, partial completion, and unintended side effects.

Third, after the automated tests pass, human developers execute every benchmark task end to end. They verify that the instruction reflects a realistic and unambiguous user need, that reasonable completion paths and valid alternative outcomes are covered, and that the interaction outcomes, terminal state, and programmatic Judge decision are consistent. Any ambiguity, boundary-condition error, missing valid outcome, or semantic mismatch triggers another Code-Agent revision followed by repeated automated testing and manual validation. Validated Apps and tasks then directly reuse the platform interfaces for state injection, reset, snapshotting, cloning, parallel rollouts, evaluation, and programmatic RL rewards.

\paragraph{Development effort.}
According to our development records, implementing a typical everyday App and its major functionality required approximately \textbf{3--4 person-days}, while a focused system App generally required less than one person-day. Implementing all 28 simulated Apps required approximately 60 person-days in total. These figures include functional exploration, implementation, business logic, synthetic data, debugging, and manual validation. Under the mature task framework, one developer assisted by Coding Agents recently added \textbf{30 tasks in approximately 8 person-hours}. The measurement covers task design, task and Judge implementation, automated testing, human review, revision, and manual end-to-end validation; approximately 80\% of the tasks were directly usable after human review, while 20\% required additional Code-Agent iteration. These internal engineering estimates are derived from project development records. The recent 30-task measurement characterizes the current marginal-extension cost under the mature shared task abstractions, Judge components, and testing tools.

The separation of App content and logic allows benchmark data to be replaced or varied independently of App code.

\section{EFSM Formalization and Declaration Syntax}\label{app:efsm}

In \sys{}, the UI navigation of every app is formalized as an extended-finite-state-machine tuple:
\begin{equation*}
  \mathcal{M} = (S,\; \Sigma,\; \Delta,\; s_0,\; D,\; G,\; U)
\end{equation*}
where $S$ is the set of UI states (each state corresponds to a unique combination of route path + query parameters); $\Sigma$ is the input alphabet (user actions); $\Delta: S \times \Sigma \times G \to S \times U$ is the transition function with guards and update operations; $s_0$ is the initial state; $D$ is the set of application state variables; $G$ is the set of guards; and $U$ is the set of update operations on $D$. Compared with a classical FSM, the EFSM extension has three roles: (1) guards allow the same input to trigger different transitions under different data states; (2) data-driven expansion allows the state space to grow according to configuration data; and (3) compound UI states model different visual presentations (popups, drawers, tabs) that share the same route path as distinct state nodes.

\paragraph{Guard examples.}\mbox{}\par\medskip
\begin{lstlisting}[language=Java]
// URL-based "from" constraint
from: { path: '/book/:id',
        search: { modal: null } }    // modal must be absent
from: { path: '/book/:id',
        search: { tab: 'comment' } } // tab must equal 'comment'

// AppState-based conditional branch
cases: [
  { to: '/user/:mid',
    search: { panel: 'recommend' },
    when: { op: 'eq',
            left: { ref: 'appState',
                    key: 'isFollowing' },
            right: false } },
  { to: '/user/:mid',
    search: { menu: 'unfollow' },
    when: { op: 'always' } }, // fallback
]

// Data-driven entry-visibility condition
ui: { condition: {
  op: 'memberOf',
  ref: 'initialShelf',
  param: 'bookId' } }
\end{lstlisting}

\paragraph{Runtime interface and DOM tagging.}\mbox{}\par\medskip
\begin{lstlisting}[language=Java]
// Hook
const { go, back } = useAppNavigate();
go('book.modal.open', { bookId: '60' });
back();

// DOM tagging (the agent does not use it directly)
<button data-trigger="book.modal.open"
        data-trigger-params='{"bookId":"60"}'>
  Open Book
</button>
\end{lstlisting}

The declarative navigation specification is both \textbf{executable and analyzable}: at runtime it drives UI navigation (a transition is fired through \texttt{go(transitionId, params)}); statically it supports consistency checking, BFS path enumeration (for candidate trajectory generation and shortest-path enumeration), and navigation-graph construction.

\section{Full Action Space}\label{app:action-space}

\begin{table}[!htbp]
  \centering\small
  \setlength{\tabcolsep}{3pt}
  \resizebox{\columnwidth}{!}{%
  \begin{tabular}{llp{4.4cm}}
    \toprule
    \textbf{Category} & \textbf{Action} & \textbf{Parameters \& description} \\
    \midrule
    \multirow{6}{*}{Physical touch}
      & \texttt{CLICK}       & \texttt{point=[x,y]}, single tap \\
      & \texttt{DOUBLE\_TAP} & \texttt{point=[x,y]}, double tap \\
      & \texttt{LONG\_PRESS} & long press to trigger context menu \\
      & \texttt{TYPE}        & \texttt{value=str} (optional point/clear), supports Chinese pinyin IME \\
      & \texttt{SWIPE}       & \texttt{point1, point2}, with inertia \\
      & \texttt{DRAG}        & \texttt{point1, point2}, no inertia \\
    \midrule
    \multirow{4}{*}{System keys}
      & \texttt{BACK}    & invoke the BackDispatcher priority chain \\
      & \texttt{HOME}    & return to desktop \\
      & \texttt{RECENT}  & open recent-tasks list \\
      & \texttt{ENTER}   & fire the Enter key \\
    \midrule
    \multirow{2}{*}{Control}
      & \texttt{WAIT}    & \texttt{value=seconds} \\
      & \texttt{AWAKE}   & \texttt{value=app\_id}, launch an app \\
    \midrule
    \multirow{3}{*}{Termination/answer}
      & \texttt{ANSWER}   & \texttt{value=str}, submit an answer (does not terminate) \\
      & \texttt{COMPLETE} & terminate the episode, declaring success \\
      & \texttt{ABORT}    & terminate the episode, declaring inability to complete \\
    \midrule
    \multirow{2}{*}{Other}
      & \texttt{INFO}     & \texttt{value=str}, ask the user a clarifying question (does not terminate) \\
      & \texttt{NOOP}     & no-op (used internally by the agent) \\
    \bottomrule
  \end{tabular}%
  }
  \caption{\sys{} action space (the Benchmark-layer \texttt{Action} abstraction, 17 in total)}
  \label{tab:action-space}
\end{table}

General-purpose models under the \texttt{generic\_v2} agent template use the full action space above (excluding \texttt{NOOP} and \texttt{INFO}, which are reserved for agent-internal use); GUI-specialized models (AutoGLM-Phone-9B, UI-TARS-1.5-7B, UI-Venus-1.5-8B, GUI-Owl-1.5-8B-Think, Step-GUI-4B) use their own native action spaces, and the agent adapter maps them to the \sys{} environment-layer \texttt{Action} abstraction. Coordinates are uniformly normalized to $[0,1000] \times [0,1000]$.

\section{App Coverage and Representative Tasks}\label{app:apps}

\begin{table}[!htbp]
  \centering\small
  \setlength{\tabcolsep}{3pt}
  \begin{tabular}{lp{2.7cm}c}
    \toprule
    \textbf{Type / Category} & \textbf{Apps} & \textbf{\#Routes} \\
    \midrule
    \multicolumn{3}{l}{\textit{Everyday apps (12)}} \\
    Social/Comm.    & WeChat       & 100 \\
                    & RedNote      & 41 \\
    Finance         & Alipay       & 48 \\
    Video/Ent.      & Bilibili     & 52 \\
    Travel          & Maps, 12306  & 17, 48 \\
    Reading/Music   & WeChat Reading, Spotify & 30, 35 \\
    Social media    & Reddit, X (Twitter) & 17, 33 \\
    Business/Prod.  & Tencent Meeting, eBay & 18, 9 \\
    \midrule
    \multicolumn{3}{l}{\textit{System apps (16)}} \\
    Launcher        & Launcher     & $-^\ast$ \\
    Core utilities  & Settings, Contacts, SMS & 138$^\dagger$, 39$^\dagger$, 8 \\
    Productivity    & Calendar, Notes, Calculator, Calculator (AOSP), AnswerSheet & 10, 7, 1, 1, $-^\ast$ \\
    System apps     & Browser, File Manager, Clock, Theme Store & 2, 7, 5, 2 \\
    Info/Nav        & Weather, Compass, Gallery & 9, 4, 5 \\
    \bottomrule
  \end{tabular}
  \caption{Simulated app coverage (12 everyday + 16 system = 28 apps). \textbf{\#Routes} counts route objects declared in each app's \texttt{navigation.declaration.ts}; compound \texttt{uiStates} (popups, drawers, tabs sharing a route path) and runtime modals are \textit{not} counted as separate routes, so the figure underestimates the number of distinguishable UI states. $^\ast$Launcher and AnswerSheet are single-screen apps without a navigation declaration. $^\dagger$Settings and Contacts use the Android AOSP data-driven preference pattern: a single \texttt{/page/:pageId} route mounts content from a page registry, so the React-level route count alone severely under-represents user-visible screens. Their figures add the reachable preference pages: Settings = 3 routes $+$ 135 reachable pages out of 623 defined (89 with interactive controls; 46 are read-only placeholder screens); Contacts = 10 routes $+$ 29 phone-preference pages (28 of them interactive).}
  \label{tab:apps}
\end{table}

Every app is populated with public or synthetic data that mimics the structure and content style of real platforms, loaded from a configurable \texttt{defaults.json}. Across the 12 everyday apps, the released world data contains over 190K entities (over 350K structured records including auxiliary indices and relations), making feeds, search results, comments, products, maps, and travel pages information-dense enough to support parameterized search, query, and deep-dive tasks.

\begin{table*}[!t]
  \centering\small
  \begin{tabular}{p{1.4cm}p{1.7cm}p{0.6cm}p{2.2cm}p{7.2cm}}
    \toprule
    \textbf{Objective} & \textbf{Composition} & \textbf{Diff.} & \textbf{App} & \textbf{Instruction example} \\
    \midrule
    operate & atomic     & L1 & Clock                & Turn on the 7:30 alarm for me \\
    query   & atomic     & L1 & Weather              & Tell me what the temperature is in Beijing right now \\
    operate & sequential & L2 & Spotify              & Add ``Shape of You'' to my Liked Songs in Spotify \\
    query   & deep\_dive & L3 & Alipay               & In the Alipay bill, accumulate the amounts and tell me which counterparty has the largest total \\
    hybrid  & transfer   & L3 & RedNote\newline $\to$ WeChat & Search ``camping'' in RedNote and send the title of the first note to Xiaohong via WeChat \\
    query   & sequential & L4 & eBay                 & I want to buy a brand-new Sony Bluetooth headset shipped from Japan; which one is the cheapest, and how much is it including shipping? \\
    hybrid  & deep\_dive & L4 & eBay+Alipay\newline $\to$ Notes & Find the cheapest brand-new AirPods on eBay, see how much my Alipay balance would have left after buying it, and write the product name, price, and remaining balance to Notes \\
    \bottomrule
  \end{tabular}
  \caption{Representative task examples}
  \label{tab:task-examples}
\end{table*}

\section{Detailed Task Taxonomy}\label{app:taxonomy}

\begin{table}[!htbp]
  \centering\small
  \setlength{\tabcolsep}{3pt}
  \begin{tabular}{lcc}
    \toprule
    \textbf{Scope} & \textbf{Test (256)} & \textbf{Train (160)} \\
    \midrule
    S1 (Single-app)      & 163 (64\%)   & 141 (88.1\%) \\
    S2 (Cross-app, 2 apps)  & 65 (25\%)    & 17 (10.6\%)  \\
    S3 (Multi-app, 3+ apps) & 28 (11\%)    & 2 (1.3\%)      \\
    \midrule
    \textbf{Total}     & \textbf{256} & \textbf{160} \\
    \bottomrule
  \end{tabular}
  \caption{Composition of Test256 / Train160 (by Scope)}
  \label{tab:task-composition}
\end{table}

Test and Train are strictly disjoint. The Train set is mostly composed of single-app tasks that cover the core skills of the 12 everyday apps; 36\% of the Test set consists of cross-app tasks, which extends beyond the training distribution and supports OOD-generalization diagnostics for cross-app performance. All tasks support parameter sampling; L3/L4 tasks additionally provide 2--3 instruction variants, which combine orthogonally with parameters to further increase diversity.

\paragraph{Test256 distribution by dimension.} Difficulty: L1=20, L2=73, L3=83, L4=80. Objective: operate=170, query=48, hybrid=38. Composition: atomic=22, sequential=110, transfer=56, deep\_dive=68.

\paragraph{Capability tags.} Each task carries 1--4 tags from the following 13-tag vocabulary:

\begin{itemize}
  \item \texttt{nav} --- navigate to a specific page or screen.
  \item \texttt{settings} --- modify app or system settings.
  \item \texttt{search} --- locate content through explicit search, filtering, or sorting mechanisms.
  \item \texttt{create} --- create new content (post, message, order, etc.).
  \item \texttt{edit} --- modify existing content or records.
  \item \texttt{delete} --- remove content, records, or accounts.
  \item \texttt{social} --- interact with other users (follow, like, comment, etc.).
  \item \texttt{extract} --- extract and report specific information from the UI.
  \item \texttt{handoff} --- transfer context or data across apps.
  \item \texttt{finance} --- perform financial operations (payment, transfer, etc.).
  \item \texttt{reasoning} --- require multi-step inference or comparison.
  \item \texttt{explore} --- locate content by navigating feeds, comment threads, long lists, or unknown page hierarchies without an explicit search entry point.
  \item \texttt{image} --- require understanding image content to complete the task.
\end{itemize}

\section{AnswerSheet Protocol Design Details}\label{app:answersheet}

\subsection{Field types and matchers}

The AnswerSheet, as a system app, provides an answer form. Each field declares a type and matcher:

\begin{itemize}
  \item \texttt{choice} field—choose from enumerated options, paired with the \texttt{exact} matcher.
  \item \texttt{number} field—numeric input, paired with the \texttt{number} matcher (with floating-point tolerance).
  \item \texttt{text} field—text input, paired with \texttt{exact} / \texttt{date} / \texttt{time} / \texttt{duration} matchers.
  \item \texttt{repeatable} multi-value list—supports scenarios that require listing multiple answers.
\end{itemize}

\subsection{Design motivation}

\textbf{Eliminating natural-language false negatives.} ``$34^\circ$C'' / ``34 degrees'' / ``about 34 Celsius'' all map to the same numeric value \texttt{34} under a \texttt{number} field. The judge can therefore perform a floating-point comparison with tolerance, without relying on preset string-normalization rules.

\textbf{Eliminating false positives from enumeration / mixed-in thinking.} A \texttt{number} field can hold only one numeric value, and a \texttt{choice} field can hold only one enumerated value. Typing ``34'' and typing ``33 or 34'' produce \textbf{physically different states}. The agent must navigate to the app, locate the field, and enter the value; each step is observable at the state layer.

\textbf{Compatibility with small models.} Some small GUI agents (e.g.\ AutoGLM-Phone-9B) often place the entire think trace in the \texttt{<answer>} field and may not stably emit purely structured answer text. Requiring them to ``fill the answer'' through GUI operations better matches their interaction-oriented training distribution.

\subsection{Format expectations and hints}

Each field carries a \texttt{hint} string, which is shown to the agent in the UI as a placeholder and explicitly indicates the expected input form, e.g.\ ``Temperature (Celsius, integer)'', ``Date (YYYY-MM-DD)'', ``Amount (CNY, two decimal places)''. The task author is responsible for providing an unambiguous format constraint in the hint. If the model still writes ``34$^\circ$C'' into a \texttt{number} field, or writes the date as ``tomorrow'' instead of a concrete date, the judge marks it wrong. This failure reflects insufficient ability to follow the output format rather than ambiguity in the evaluation design.

\subsection{Compensation for execution cost}

The AnswerSheet introduces an additional ``switch app + fill form'' GUI workflow for query tasks. To compensate for this execution cost, tasks with \texttt{answer\_fields} are \textbf{given an additional 15-step budget} (i.e.\ L1: $15+15=30$, L4: $60+15=75$), corresponding to a reasonable number of actions for ``switching to AnswerSheet + filling fields + submitting''.

\section{Detailed Experimental Configuration}\label{app:exp-config}

\textbf{Inference configuration.} General-purpose models (\texttt{generic\_v2}) uniformly use the following settings: decoding \texttt{temperature}=0.1, \texttt{top\_p}=0.95, \texttt{frequency\_penalty}=0, \texttt{max\_tokens}=4096; single-step LLM call timeout 300\,s; a 0.8\,s wait after each action to allow the UI render to stabilize; and \texttt{loop\_detect}=10 (early termination when the same action is repeated $\geq$10 times in a row). The screenshot is provided at full resolution ($1080 \times 2400$ physical $\to$ 0--1000 normalized coordinates), and the dialogue history is managed in a ``current step carries the screenshot + earlier steps keep only the LLM text response'' format to avoid context growth over long episodes. The prompt templates and decoding configurations of GUI-specialized models follow their original papers / official implementations, and the execution layer shares the environment constraints above.

\textbf{GRPO training configuration.} The Sim-to-Real training run uses Qwen3-VL-4B-Instruct as the initial policy, 3 GPUs on a single node, 96 parallel environment instances. Rollouts use \texttt{vLLM} asynchronous mode with group size $k=8$, train batch size 12, PPO mini-batch size 12, and per-GPU micro-batch size 2. The optimizer learning rate is $10^{-6}$, gradient clipping is 1.0, $\gamma=1.0$, $\lambda=1.0$, KL coefficient is 0.01, and the asymmetric clipping range is 0.2/0.28. We use maximum prompt length 32768, maximum response length 1024, rollout maximum model length 40960, and training-time agent decoding temperature 0.7 (validation temperature 0.1). The environment pool uses page-level isolation, a 0.8\,s delay after each action.

\textbf{Reward function.} The training reward is computed from structured rollout artifacts. Let $p \in [0,1]$ denote task progress, i.e., the fraction of goal checks passed. The base reward is $r=p$. For AnswerSheet tasks, if the agent submits the sheet but any answer field is wrong, the reward is recomputed after removing the bookkeeping check \texttt{answer\_sheet.submitted}, so submitting an incorrect sheet does not provide extra progress credit. Multiplicative discounts are then applied:
\begin{equation*}
\begin{aligned}
r \leftarrow p^\prime
  &\cdot 0.8^{\mathbb{I}[\text{goal success} \wedge \neg \text{clean}]} \\
  &\cdot 0.8^{\mathbb{I}[\text{false complete} \wedge p^\prime>0]} \\
  &\cdot 0.5^{\mathbb{I}[\text{post-success abort}]}
   \cdot 0.5^{\mathbb{I}[\text{overdue}]},
\end{aligned}
\end{equation*}
where $p^\prime$ is either the original progress $p$ or the AnswerSheet-adjusted progress described above. \textit{Goal success} means the task goal state is reached, \textit{clean} means no unexpected state changes are detected, and \textit{false complete} means the agent terminates with \texttt{COMPLETE} but the episode is not a full success. The final two terms penalize cases where the goal state is reached but the agent does not correctly declare completion: \textit{post-success abort} means it terminates with \texttt{ABORT}, while \textit{overdue} means it keeps acting until truncation. Binary task correctness for reporting is still the simulator's final success signal, not the shaped reward.

\textbf{Notes on the evaluated models.}
\begin{itemize}
  \item Gemini 3.1 Pro~\cite{Gemini31Pro}: a reasoning model from Google DeepMind.
  \item Doubao-Seed-2.0-Pro~\cite{DoubaoSeed20Pro}: a multimodal model from ByteDance Seed.
  \item Qwen3.6-Plus~\cite{Qwen36Plus}: a multimodal model from Alibaba Tongyi Qianwen.
  \item AutoGLM-Phone-9B~\cite{AutoGLM}: a mobile-oriented GUI agent from Zhipu AI.
  \item UI-TARS-1.5-7B~\cite{UITARS}: a GUI agent from ByteDance Seed.
  \item UI-Venus-1.5-8B~\cite{UIVenus15}: full-trajectory online RL training; AndroidWorld SOTA.
  \item GUI-Owl-1.5-8B-Think~\cite{GUIOw15}: introduced in Mobile-Agent-v3.5; multi-platform RL with MRPO.
  \item Step-GUI-4B~\cite{StepGUI}: Calibrated Step Reward System.
  \item Qwen3-VL-4B-Instruct~\cite{Qwen3VL}: a vision-language model from Qwen, used as the base model for our Sim-to-Real training experiments.
\end{itemize}

\section{Full Result Decomposition}\label{app:results-detail}

\subsection{SR by taxonomy dimension}

Table~\ref{tab:results-by-dim} decomposes the test-set Success Rate along all three taxonomy axes (Difficulty, Objective, Composition; §\ref{sec:bench:taxonomy}), extending the difficulty-only breakdown in Table~\ref{tab:main-results}.

\begin{table*}[!t]
  \centering\small
  \setlength{\tabcolsep}{4pt}
  \resizebox{\textwidth}{!}{%
  \begin{tabular}{l|cccc|ccc|cccc}
    \toprule
    & \multicolumn{4}{c|}{\textbf{Difficulty}} & \multicolumn{3}{c|}{\textbf{Objective}} & \multicolumn{4}{c}{\textbf{Composition}} \\
    \textbf{Model} & L1 (20) & L2 (73) & L3 (83) & L4 (80) & oper. (170) & query (48) & hybrid (38) & atom. (22) & seq. (110) & trans. (56) & deep (68) \\
    \midrule
    \rowcolor{gray!8} \multicolumn{12}{l}{\textit{Proprietary models}} \\
    Gemini 3.1 Pro       & 97.5 & 83.6 & 63.3 & 21.9 & 56.8 & 64.6 & 60.5 & 81.8 & 76.8 & 33.0 & 43.4 \\
    Doubao-Seed-2.0-Pro  &100.0 & 93.2 & 48.2 &  6.2 & 52.4 & 52.1 & 50.0 & 77.3 & 70.0 & 25.0 & 36.8 \\
    Qwen3.6-Plus         &100.0 & 78.1 & 44.6 &  3.8 & 41.2 & 50.0 & 60.5 & 81.8 & 61.8 & 16.1 & 32.4 \\
    \midrule
    \rowcolor{gray!8} \multicolumn{12}{l}{\textit{Open-source GUI-specialized models}} \\
    AutoGLM-Phone-9B     & 86.2 & 33.6 &  9.6 &  1.9 & 20.7 & 23.4 & 12.5 & 56.8 & 25.7 &  7.1 &  9.6 \\
    UI-TARS-1.5-7B       & 77.5 & 21.9 &  3.0 &  1.6 & 14.0 & 20.3 &  4.6 & 37.5 & 20.2 &  2.7 &  4.8 \\
    UI-Venus-1.5-8B      & 85.0 & 21.9 &  6.0 &  1.9 & 16.9 & 15.6 &  8.6 & 37.5 & 20.7 &  6.2 &  7.4 \\
    GUI-Owl-1.5-8B-Think & 76.2 & 26.0 &  4.2 &  1.2 & 16.9 & 14.6 &  7.9 & 44.3 & 20.9 &  2.7 &  6.6 \\
    Step-GUI-4B          & 83.8 & 17.8 &  2.4 &  1.6 & 16.3 &  7.8 &  3.9 & 30.7 & 21.4 &  0.4 &  3.7 \\
    \midrule
    \rowcolor{gray!8} \multicolumn{12}{l}{\textit{Open-source generalist models}} \\
    Qwen3-VL-4B-Instruct          & 71.2 & 12.3 &  0.6 &  0.3 & 12.9 &  3.6 &  0.7 & 25.0 & 15.5 &  0.9 &  1.5 \\
    \bottomrule
  \end{tabular}%
  }
  \caption{Success Rate (\%) decomposed by Difficulty / Objective / Composition}
  \label{tab:results-by-dim}
\end{table*}

\subsection{Trajectory-length diagnostics}

Table~\ref{tab:trajectory-length} reports the mean episode length (\textbf{Steps}) and the mean length of successful trajectories (\textbf{Steps\checkmark}) on the \bench{} test set. Successful trajectories cluster around 8--14 steps across models, while overall episode length varies substantially, reflecting how often weaker models exhaust the step budget without succeeding.

\begin{table}[!htbp]
  \centering
  \begingroup
  \footnotesize
  \setlength{\tabcolsep}{4pt}
  \begin{tabular*}{\columnwidth}{@{}l@{\extracolsep{\fill}}cc@{}}
    \toprule
    \textbf{Model} & \textbf{Steps} & \textbf{Steps\checkmark} \\
    \midrule
    \rowcolor{gray!8} \multicolumn{3}{l}{\textit{Proprietary models}} \\
    Gemini 3.1 Pro       & 16.4 & 13.4 \\
    Doubao-Seed-2.0-Pro  & 17.1 & 12.8 \\
    Qwen3.6-Plus         & 20.6 & 14.0 \\
    \midrule
    \rowcolor{gray!8} \multicolumn{3}{l}{\textit{Open-source GUI-specialized models}} \\
    AutoGLM-Phone-9B     & 26.8 & 13.1 \\
    UI-TARS-1.5-7B       & 27.6 & 13.0 \\
    UI-Venus-1.5-8B      & 21.6 & 11.0 \\
    GUI-Owl-1.5-8B-Think & 24.9 & 11.7 \\
    Step-GUI-4B          & 31.7 & 11.8 \\
    \midrule
    \rowcolor{gray!8} \multicolumn{3}{l}{\textit{Open-source generalist models}} \\
    Qwen3-VL-4B-Instruct          & 17.6 &  7.9 \\
    \bottomrule
  \end{tabular*}
  \endgroup
  \caption{Trajectory-length diagnostics on the \bench{} test set. \textbf{Steps}=mean episode length; \textbf{Steps\checkmark}=mean length of successful trajectories.}
  \label{tab:trajectory-length}
\end{table}

\subsection{High-Risk subset}

\begin{table}[!htbp]
  \centering\small
  \setlength{\tabcolsep}{4pt}
  \resizebox{\columnwidth}{!}{%
  \begin{tabular}{lccc}
    \toprule
    \textbf{Model} & \textbf{Payment (7)} & \textbf{T256-Risk (7)} & \textbf{All (14)} \\
    \midrule
    \rowcolor{gray!8} \multicolumn{4}{l}{\textit{Proprietary models}} \\
    Gemini 3.1 Pro       & 64.3 & 71.4 & 67.9 \\
    Doubao-Seed-2.0-Pro  &  0.0 & 71.4 & 35.7 \\
    Qwen3.6-Plus         & 28.6 & 42.9 & 35.7 \\
    \midrule
    \rowcolor{gray!8} \multicolumn{4}{l}{\textit{Open-source GUI-specialized models}} \\
    AutoGLM-Phone-9B     &  3.6 & 28.6 & 16.1 \\
    UI-TARS-1.5-7B       &  0.0 & 25.0 & 12.5 \\
    UI-Venus-1.5-8B      & 10.7 & 14.3 & 12.5 \\
    GUI-Owl-1.5-8B-Think &  3.6 & 25.0 & 14.3 \\
    Step-GUI-4B          &  0.0 & 14.3 &  7.1 \\
    \midrule
    \rowcolor{gray!8} \multicolumn{4}{l}{\textit{Open-source generalist models}} \\
    Qwen3-VL-4B-Instruct          &  0.0 & 28.6 & 14.3 \\
    \bottomrule
  \end{tabular}%
  }
  \caption{High-Risk subset SR (\%): financial operations / account-credential modifications / irreversible deletions}
  \label{tab:high-risk-results}
\end{table}

The High-Risk subset comprises 7 standalone \texttt{Payment} tasks (money transfer, card binding, subscription renewal, etc.) plus 7 high-risk tasks within test256 (e.g., account registration, account deactivation, large transfers, and message/data deletion), 14 in total. This subset characterizes execution capability in \textbf{irreversible / high-consequence} scenarios; it differs from a safety evaluation that tests refusal of harmful or inappropriate instructions. The numbers in this table are completion success rates and do not measure whether the model should refuse the operation. See the Ethical Considerations section.

\subsection{Sim-to-Real simulation-side breakdown by difficulty}

\begin{table}[!htbp]
  \centering\small
  \setlength{\tabcolsep}{3pt}
  \resizebox{\columnwidth}{!}{%
  \begin{tabular}{lcccc|c}
    \toprule
    \textbf{Model} & \textbf{L1 (20)} & \textbf{L2 (73)} & \textbf{L3 (83)} & \textbf{L4 (80)} & \textbf{All} \\
    \midrule
    Qwen3-VL-4B-Instruct (base)             & 71.2 & 12.3 &  0.6 & 0.3 &  9.4 \scriptsize{$\pm$0.6} \\
    Qwen3-VL-4B-10s (trained)      & 92.5 & 37.7 & 11.7 & 1.2 & 22.2 \scriptsize{$\pm$1.2} \\
    \midrule
    \textbf{$\Delta$ (training gain, pt)} & \textbf{+21.3} & \textbf{+25.4} & \textbf{+11.1} & \textbf{+0.9} & \textbf{+12.8} \\
    \bottomrule
  \end{tabular}%
  }
  \caption{Sim-to-Real training gain on the simulation side, broken down by difficulty (256-task test set)}
  \label{tab:sim2real-by-level}
\end{table}

\subsection{Sim-to-Real outcome-stratified task sampling}\label{app:sim2real-buckets}

Tasks are bucketed by the number of passes out of 4 base / 4 trained rollouts: \textbf{Uplift} (base $\leq$1/4, trained $\geq$3/4, all 26 selected); \textbf{Stable-pass} (both $\geq$3/4, all 21 selected); \textbf{Mid} (partial uplift, all 20 selected); \textbf{Regression} (base $\geq$3/4, trained $\leq$1/4, 0 instances, suggesting that no severe regression is observed under this sampling protocol); and \textbf{Stable-fail} (both $\leq$1/4, 189 tasks). From the three signal buckets we select all 67 tasks (uplift 26 + stable-pass 21 + mid 20); from Stable-fail we additionally sample 15 tasks at random as a sanity check, re-sampling any task that cannot be equivalently reproduced on the real device.

Of the 67 signal-bucket tasks, 8 are dropped because they cannot be equivalently reproduced on the real device: 3 irreversible account-level modifications, 1 non-reversible consumption-style operation, and 4 tasks that require preset states the real device cannot equivalently reproduce (synthetic meeting histories, preset message sessions, etc.). The final \textbf{59 signal-bucket tasks} are the headline subset; combined with the 15 stable-fail sanity-check tasks, 74 tasks are run on the real device in total. These 8 unrun tasks illustrate the complementary value of the simulator: within \sys{}, they can be configured to arbitrary initial states and rolled back without real-world consequences, while on a real device such configurations are either not equivalently reproducible or require prohibitive cost.

\subsection{Same-outcome Trajectory-Length Breakdown}\label{app:fidelity-breakdown}

Table~\ref{tab:fidelity-breakdown} compares trajectory length only on same-outcome pairs: real-device successes are paired with simulator successful rollouts for the same task, and real-device failures are paired with simulator failed rollouts for the same task.

\begin{table}[!htbp]
  \centering\small
  \setlength{\tabcolsep}{3pt}
  \resizebox{\columnwidth}{!}{%
  \begin{tabular}{llccccc}
    \toprule
    \textbf{Model} & \textbf{Slice} & \textbf{Pairs} & \textbf{Sim} & \textbf{Real} & \textbf{$\Delta$} & \textbf{MAE} \\
    \midrule
    \multirow{4}{*}{Trained}
      & \textbf{operate-succ.} & \textbf{91} & \textbf{10.08} & \textbf{12.20} & \textbf{+2.12} & \textbf{5.40} \\
      & operate-fail           & 12 & 23.75 & 17.42 & $-6.33$ & 8.50 \\
      & query-succ.            & 28 & 13.71 & 5.54  & $-8.18^\star$ & 8.18 \\
      & hybrid-succ.           & 17 & 18.06 & 15.06 & $-3.00^\star$ & 3.94 \\
    \midrule
    \multirow{5}{*}{Base}
      & \textbf{operate-succ.} & \textbf{38} & \textbf{5.00} & \textbf{6.03} & \textbf{+1.03} & \textbf{1.34} \\
      & operate-fail           & 57 & 16.74 & 38.32 & $+21.58^\dagger$ & 21.75 \\
      & query-succ.            & 6  & 15.50 & 6.00  & $-9.50^\star$ & 9.50 \\
      & query-fail             & 28 & 16.21 & 30.57 & $+14.36^\dagger$ & 27.79 \\
      & hybrid-fail            & 39 & 17.64 & 34.23 & $+16.59^\dagger$ & 19.97 \\
    \bottomrule
  \end{tabular}%
  }
  \\[2pt]
  \scriptsize $^\star$ Query/hybrid success only; sim includes the AnswerSheet submission stage absent on real. $^\dagger$ Sim-side loop-detect early stop ($\geq$10 identical actions) vs.\ real-side run-to-budget on base-model flailing.
  \caption{Same-outcome paired trajectory length on sim vs.\ real, broken down by Objective and outcome. $\Delta = $ Real $-$ Sim.}
  \label{tab:fidelity-breakdown}
\end{table}

Three observations.

\textbf{(i) Operate-success length remains comparable.} On paired tasks that succeed in both environments, real-device operate trajectories are only modestly longer than simulator trajectories (trained +2.12 steps; base +1.03 steps), consistent with similar successful-operation path lengths across environments.

\textbf{(ii) Query/hybrid rows reflect protocol asymmetry.} Real-device query successes are shorter because real-device query tasks do not include the simulator-only AnswerSheet submission workflow.

\textbf{(iii) Failure rows are termination diagnostics, not trajectory-length evidence.} Simulator runs use loop-based early stopping, while the real-device runs in this study do not; consequently, many base failures run until the task step budget on the real device.

\section{Reference-Model Sensitivity of the L1--L4 Stratification}\label{app:panel-sensitivity}

The primary L1--L4 strata in §\ref{sec:bench} are calibrated by the joint SR+PR criterion over eight reference models. The criterion is applied sequentially: L1 requires mean SR $\geq75\%$ and mean PR $\geq75\%$; L2 contains remaining tasks with mean SR $\geq25\%$ and mean PR $\geq50\%$; L3 contains remaining tasks with mean SR $>0$ and mean PR $\geq25\%$; L4 contains the rest. To test sensitivity to the choice and number of reference models, we re-run the same calibration using four reference models: \{Gemini 3.1 Pro, Doubao-Seed-2.0-Pro, UI-Venus-1.5-8B, Step-GUI-4B\}. Qwen3-VL-4B-Instruct and its trained variant remain held out in both calibrations, avoiding calibration bias toward the Sim-to-Real lift analysis.

The bucket counts shift from $20/73/83/80$ under the 8-model calibration to $25/99/58/74$ under the 4-model calibration. The corresponding mean SR/PR values remain well separated: 8-model L1--L4 means are $(88.3,90.7)$, $(47.0,64.0)$, $(22.7,38.3)$, $(5.0,15.0)$, while 4-model means are $(89.0,91.2)$, $(50.4,63.3)$, $(24.6,38.5)$, $(3.5,18.4)$. Table~\ref{tab:panel-sensitivity} reports the per-model SR breakdown under both calibrations.

Two qualitative observations from §\ref{sec:exp:bench} and §\ref{sec:exp:sim2real} are robust:

\begin{enumerate}
\item \textbf{Sim-to-Real lift concentrates on L1--L2 and diminishes sharply at L3--L4 under both calibrations.} The primary 8-model calibration yields $+21.3/+25.4/+11.1/+0.9$\,pt, while the 4-model calibration yields $+23.0/+22.5/+7.3/+0.7$\,pt. In both cases, most of the training lift lies in L1--L2 and nearly vanishes on L4.
\item \textbf{L4 isolates the frontier under both calibrations.} Under the 8-model calibration, only Gemini 3.1 Pro stays meaningfully above the floor on L4 (21.9\%), while every other model is $\leq6.2\%$. Under the 4-model calibration, Gemini remains the only model above 10\% on L4 (12.2\%), while all other models are $\leq8.1\%$. The trained 4B model also exceeds AutoGLM-Phone-9B on L1--L2 under both calibrations.
\end{enumerate}

The full overall SR ($9.4\%\to 22.2\%$, $+12.8$\,pt for trained vs.\ base) is invariant by construction because the test set is fixed at 256 tasks.

\begin{table*}[!htbp]
  \centering\small
  \setlength{\tabcolsep}{4pt}
  \resizebox{\textwidth}{!}{%
  \begin{tabular}{l|cccc|cccc}
    \toprule
    & \multicolumn{4}{c|}{\textbf{8 reference models (primary)}} & \multicolumn{4}{c}{\textbf{4 reference models (sensitivity check)}} \\
    \textbf{Model} & L1 (20) & L2 (73) & L3 (83) & L4 (80) & L1 (25) & L2 (99) & L3 (58) & L4 (74) \\
    \midrule
    \rowcolor{gray!8} \multicolumn{9}{l}{\textit{Proprietary models}} \\
    Gemini 3.1 Pro       &  97.5 & 83.6 & 63.3 & 21.9 &  98.0 & 85.4 & 56.0 & 12.2 \\
    Doubao-Seed-2.0-Pro  & 100.0 & 93.2 & 48.2 &  6.2 & 100.0 & 88.9 & 34.5 &  0.0 \\
    Qwen3.6-Plus         & 100.0 & 78.1 & 44.6 &  3.8 & 100.0 & 66.7 & 34.5 &  8.1 \\
    \midrule
    \rowcolor{gray!8} \multicolumn{9}{l}{\textit{Open-source GUI-specialized models}} \\
    AutoGLM-Phone-9B     &  86.2 & 33.6 &  9.6 &  1.9 &  78.0 & 23.5 &  8.2 &  5.1 \\
    UI-TARS-1.5-7B       &  77.5 & 21.9 &  3.0 &  1.6 &  67.0 & 15.9 &  0.4 &  3.4 \\
    UI-Venus-1.5-8B      &  85.0 & 21.9 &  6.0 &  1.9 &  79.0 & 15.9 &  5.2 &  1.4 \\
    GUI-Owl-1.5-8B-Think &  76.2 & 26.0 &  4.2 &  1.2 &  64.0 & 19.2 &  5.6 &  0.7 \\
    Step-GUI-4B          &  83.8 & 17.8 &  2.4 &  1.6 &  79.0 & 11.6 &  2.6 &  0.3 \\
    \midrule
    \rowcolor{gray!8} \multicolumn{9}{l}{\textit{Open-source generalist models (Sim-to-Real subject)}} \\
    Qwen3-VL-4B-Instruct (base)         & 71.2 & 12.3 &  0.6 & 0.3 & 63.0 &  7.6 & 0.9 & 0.3 \\
    Qwen3-VL-4B-10s (trained)  & 92.5 & 37.7 & 11.7 & 1.2 & 86.0 & 30.1 & 8.2 & 1.0 \\
    \midrule
    \textbf{Trained $-$ base lift (pt)} & \textbf{+21.3} & \textbf{+25.4} & \textbf{+11.1} & \textbf{+0.9} & \textbf{+23.0} & \textbf{+22.5} & \textbf{+7.3} & \textbf{+0.7} \\
    \bottomrule
  \end{tabular}%
  }
  \caption{Reference-model sensitivity of the L1--L4 stratification under the joint SR+PR criterion. Qwen3-VL-4B-Instruct and Qwen3-VL-4B-10s are excluded from both calibrations. All SR numbers are mean across the same trial counts as Table~\ref{tab:main-results}.}
  \label{tab:panel-sensitivity}
\end{table*}

\section{Detailed VLM-Judge Misjudgment Audit}\label{app:vlm-audit}

\begin{table}[!htbp]
  \centering\small
  \setlength{\tabcolsep}{4pt}
  \resizebox{\columnwidth}{!}{%
  \begin{tabular}{lrrrr}
    \toprule
    \textbf{Model} & \textbf{FP} & \textbf{FN} & \textbf{Total} & \textbf{Misjudge} \\
    \midrule
    Base (Qwen3-VL-4B-Instruct)                    & 4 & 1 &  5/59 &  8.5\% \\
    Trained (+Sim RL, \texttt{train10s})  & 4 & 3 &  7/59 & 11.9\% \\
    \midrule
    \textbf{Total}                         & 8 & 4 & 12/118 & \textbf{10.2\%} \\
    \bottomrule
  \end{tabular}%
  }
  \caption{Manual review of real-device VLM-judge misjudgments on the 59-task signal-bucket subset (118 trajectories = 59 tasks $\times$ 2 models). The 30 stable-fail trajectories (15 tasks $\times$ 2 models) incur 0 misjudgments and are not included in the rate; including them would dilute the rate to 12/148 = 8.1\%.}
  \label{tab:vlm-judge}
\end{table}

We use Qwen3.6-Plus as the VLM judge for real-device evaluation because it is among the closed-source models with strong multimodal reasoning capability and relatively low API cost. The real-device pass/fail labels used to compute Figure~\ref{fig:sim2real} are corrected according to this manual audit. The 12 misjudgment instances cover 9 unique tasks; some tasks are misjudged for both the base and trained models. The 30 stable-fail trajectories ($n=15$ tasks $\times$ 2 models) incur 0 misjudgments because both models fail in obvious patterns that the VLM judge reads correctly. We therefore report 10.2\% on the 118-trajectory signal-bucket subset, where misjudgments by construction occur on non-trivial trajectories, as the headline rate, with the 8.1\% (12/148) on the broader pool noted for completeness. The misjudge rate of the trained model (11.9\%) is slightly higher than that of the base model (8.5\%) because the trained model produces more complex trajectories, which give the VLM more ``declarative-statement'' surface that can lead to errors. This phenomenon warrants further study.

\paragraph{Judge-model robustness check.}
To test whether the observed errors are specific to Qwen3.6-Plus, we re-judge the same saved real-device trajectories with GPT-5.4~\cite{GPT54} without re-running the agents. We use the same manually audited labels as ground truth and exclude protocol-level manual exceptions caused by real-app anomalies from the judge-error count. GPT-5.4 yields the same aggregate error rate, but with a different distribution across base and trained trajectories (Table~\ref{tab:gpt-rejudge}).

\begin{table}[!htbp]
  \centering\small
  \setlength{\tabcolsep}{4pt}
  \begin{tabular}{lrrrr}
    \toprule
    \textbf{Judge} & \textbf{Base} & \textbf{Trained} & \textbf{Total} & \textbf{Error rate} \\
    \midrule
    Qwen3.6-Plus & 5/59 & 7/59 & 12/118 & 10.2\% \\
    GPT-5.4      & 3/59 & 9/59 & 12/118 & 10.2\% \\
    \bottomrule
  \end{tabular}
  \caption{Robustness check: VLM-judge error rate when the same saved real-device trajectories are re-judged with GPT-5.4.}
  \label{tab:gpt-rejudge}
\end{table}

\section{Case Study}\label{app:case-study}

\begin{figure*}[t]
  \centering
  \begin{subfigure}[t]{0.235\textwidth}
    \includegraphics[width=\linewidth]{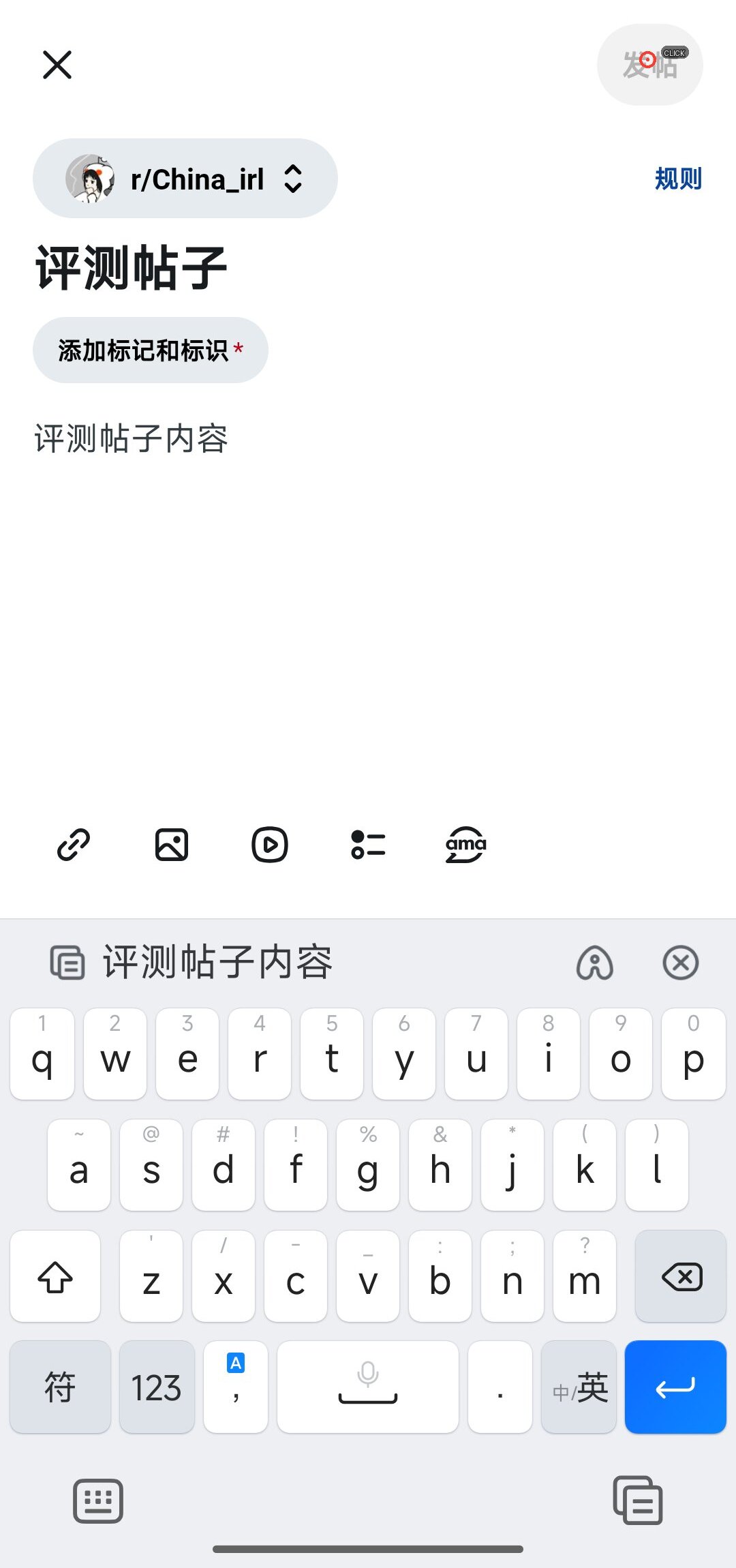}
    \caption{Step 13: clicks the grayed-out ``Post''.}
    \label{fig:reddit-tr-13}
  \end{subfigure}\hfill
  \begin{subfigure}[t]{0.235\textwidth}
    \includegraphics[width=\linewidth]{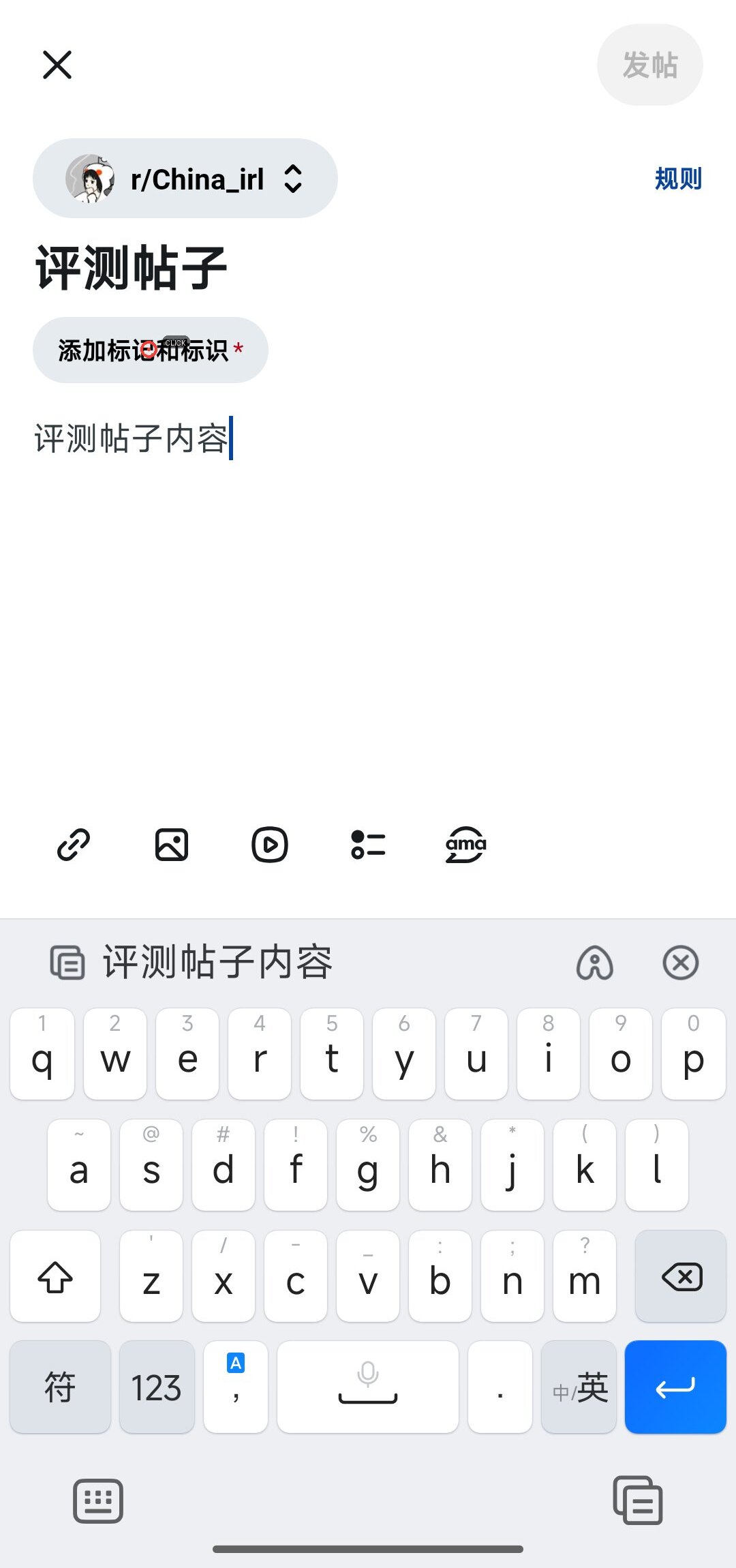}
    \caption{Step 15: notices the asterisk on the flair pill and clicks it.}
    \label{fig:reddit-tr-15}
  \end{subfigure}\hfill
  \begin{subfigure}[t]{0.235\textwidth}
    \includegraphics[width=\linewidth]{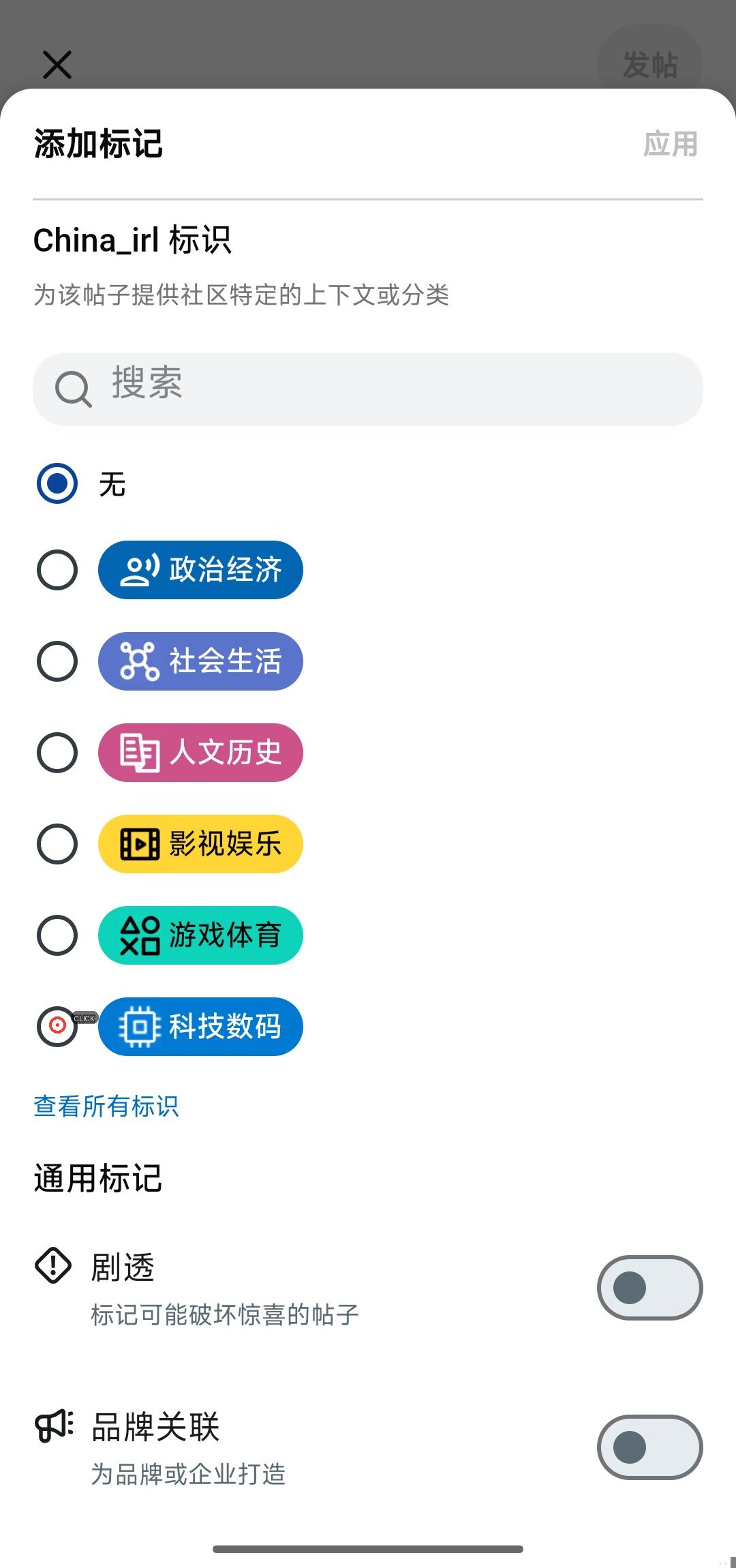}
    \caption{Step 16: picks the ``Tech'' flair.}
    \label{fig:reddit-tr-16}
  \end{subfigure}\hfill
  \begin{subfigure}[t]{0.235\textwidth}
    \includegraphics[width=\linewidth]{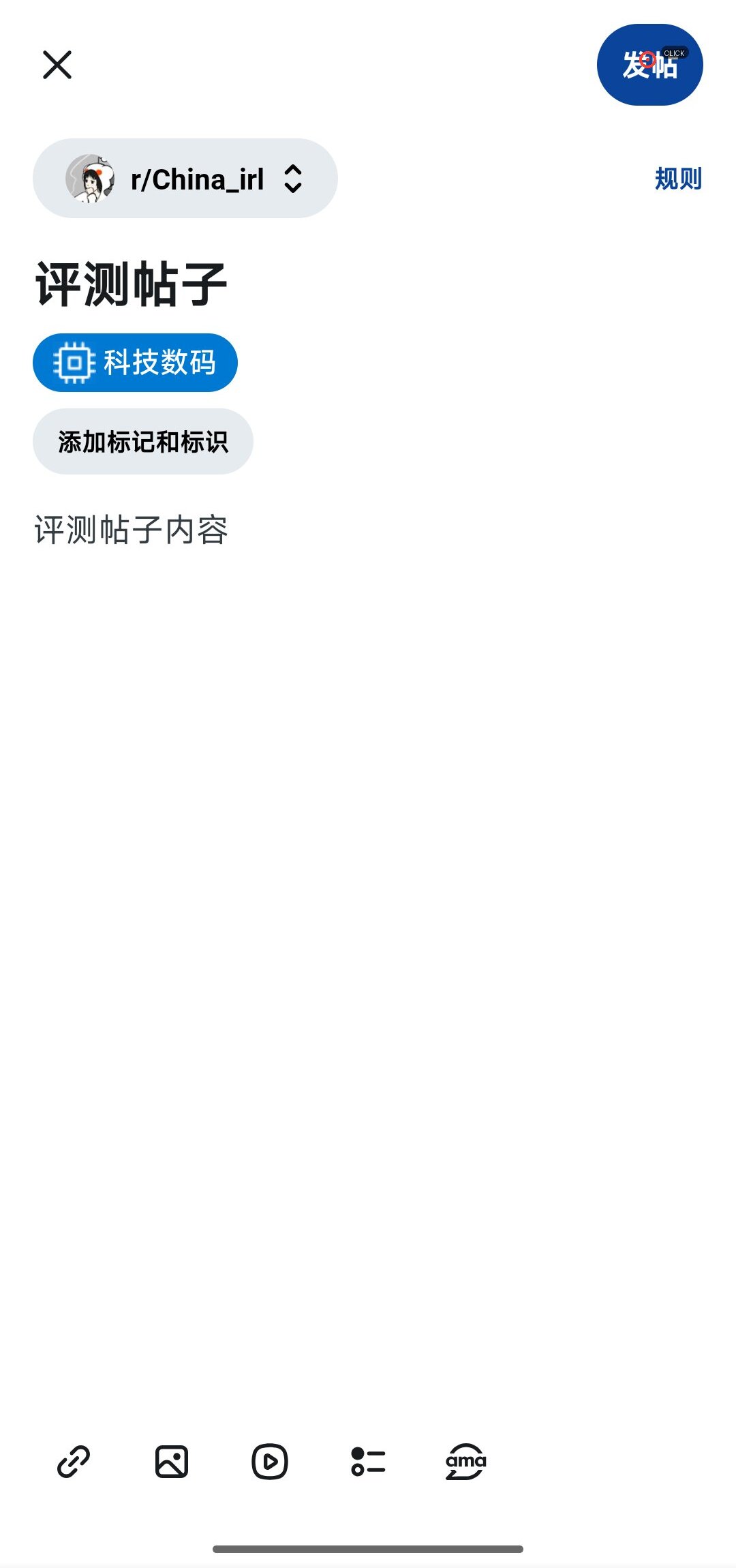}
    \caption{Step 18: ``Post'' is now blue; submits successfully.}
    \label{fig:reddit-tr-18}
  \end{subfigure}

  \medskip

  \begin{subfigure}[t]{0.235\textwidth}
    \includegraphics[width=\linewidth]{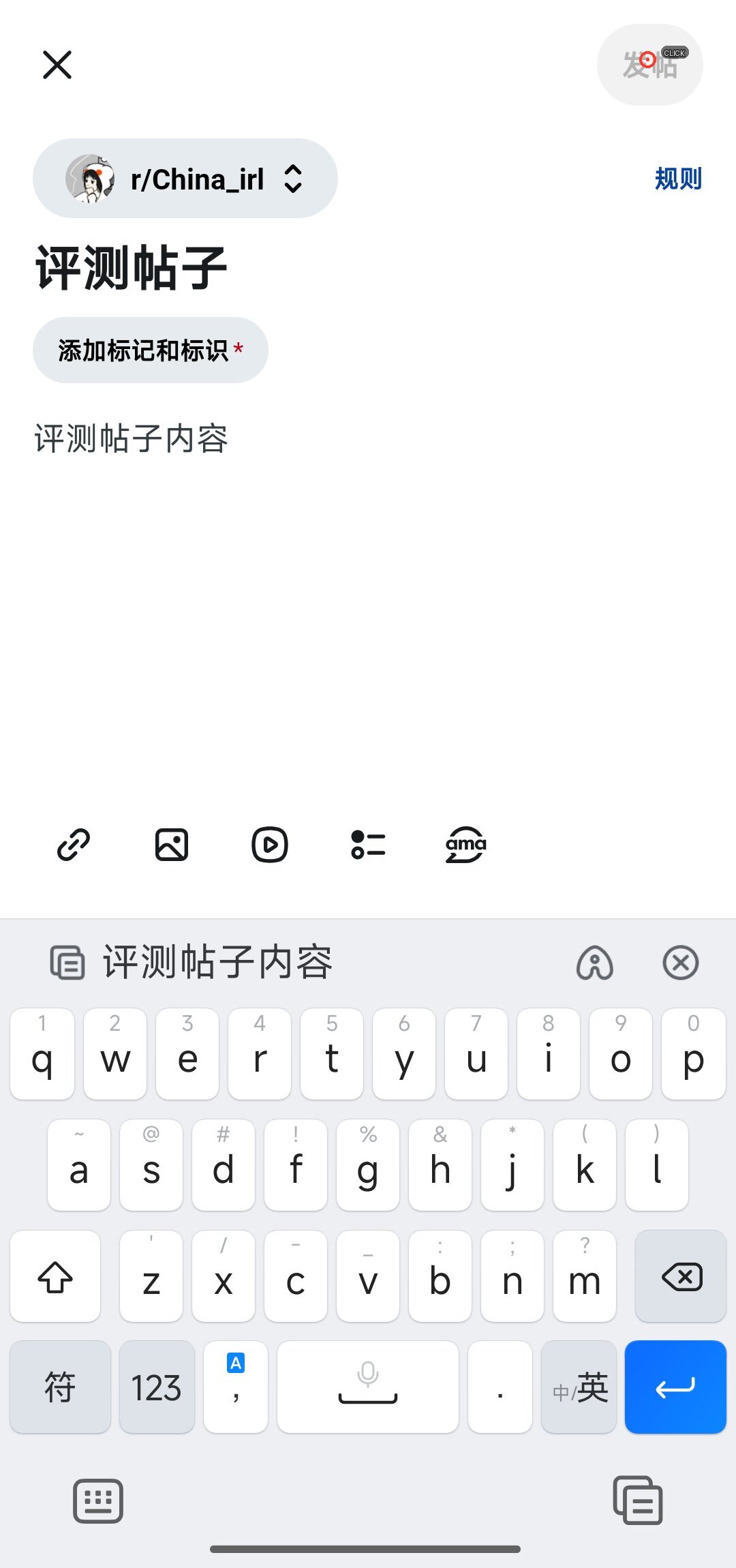}
    \caption{Step 10: first click on the grayed-out ``Post''.}
    \label{fig:reddit-ba-10}
  \end{subfigure}\hfill
  \begin{subfigure}[t]{0.235\textwidth}
    \includegraphics[width=\linewidth]{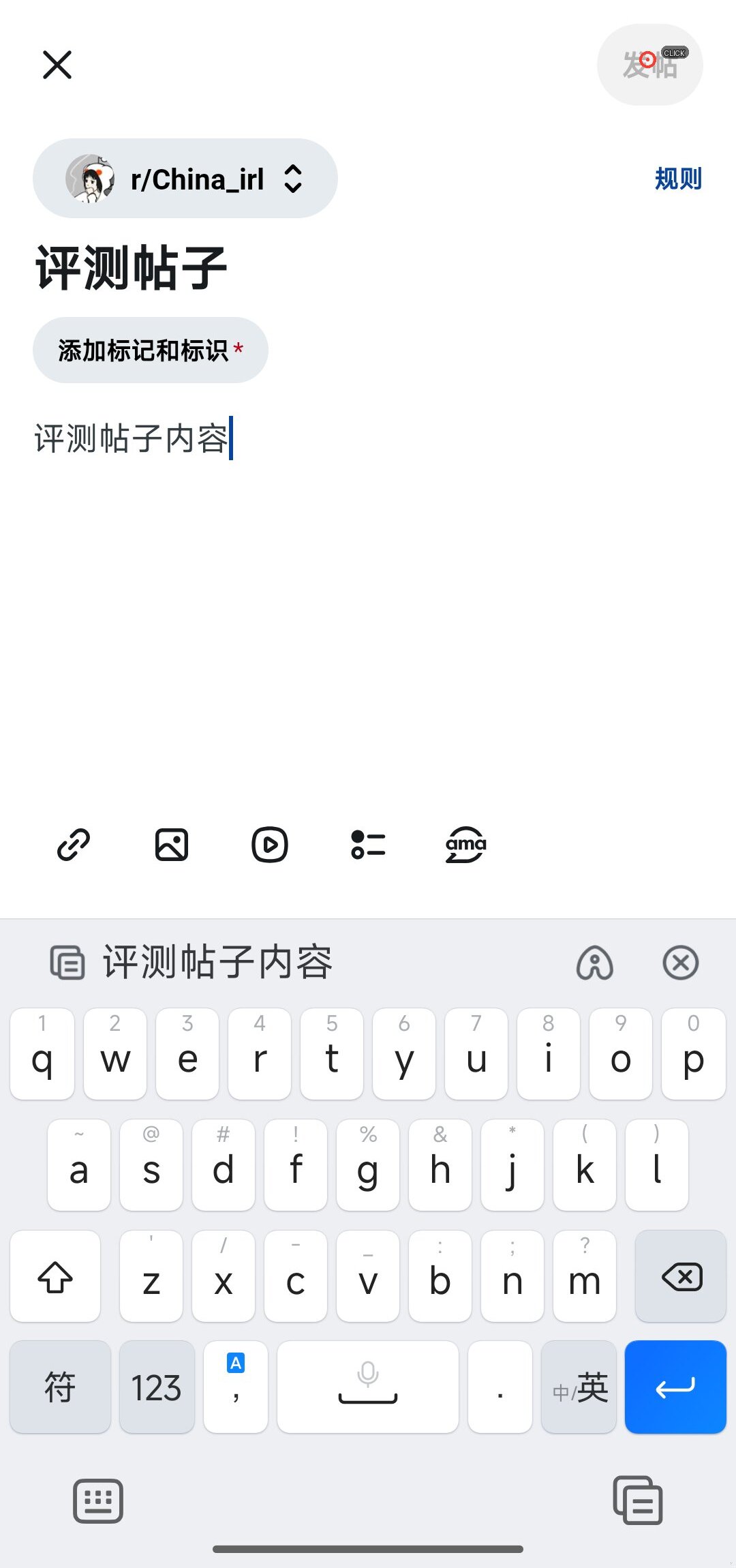}
    \caption{Step 30: still clicking; identical reasoning.}
    \label{fig:reddit-ba-30}
  \end{subfigure}\hfill
  \begin{subfigure}[t]{0.235\textwidth}
    \includegraphics[width=\linewidth]{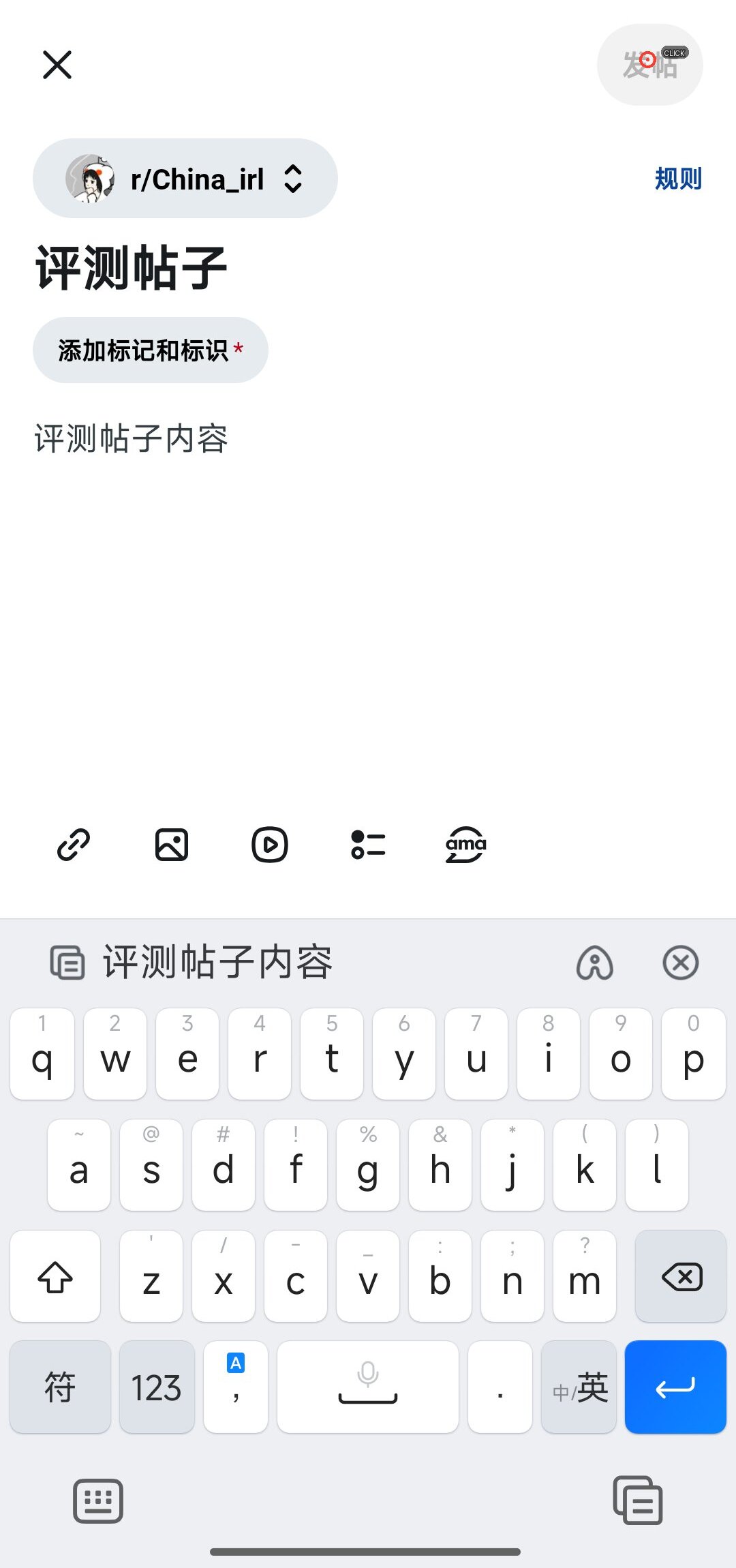}
    \caption{Step 60: 60-step budget exhausted.}
    \label{fig:reddit-ba-60}
  \end{subfigure}\hfill
  \begin{subfigure}[t]{0.235\textwidth}
    \centering\vspace{8pt}\footnotesize\itshape%
    From step 10 onward the base model produces byte-for-byte identical reasoning at every step (see Listing~\ref{lst:reddit-trained-think}, bottom block), repeatedly clicking the disabled ``Post'' until the budget expires.
  \end{subfigure}

  \caption{\textbf{Sim-to-Real OOD generalization on \texttt{Reddit\_CreatePostToCommunity}.} The real-device r/China\_irl community requires a flair tag before submission. \textbf{Top row}---trained model recovery (4 keyframes from a 22-step trajectory): step~13 clicks the grayed ``Post''; step~15 attends to the asterisk on the \texttt{Add tags \& flair} pill and infers that flair is a required field; step~16 picks the ``Tech'' flair; step~18 the ``Post'' button has turned blue and the model submits successfully. \textbf{Bottom row}---base model loop (3 keyframes spanning a 60-step trajectory): the screen and reasoning remain unchanged from step~10 onward, and the model clicks the disabled ``Post'' until the step budget is exhausted. Red \texttt{CLICK} badges mark the tap target of the model on each frame; device status bars have been cropped.}
  \label{fig:reddit-case}
\end{figure*}

\paragraph{Sim-to-Real OOD generalization: a case study.} The real-device community used in \texttt{Reddit\_CreatePostToCommunity} requires a flair tag before a post can be submitted. In a single real-device trial, the base model repeats clicks on the grayed-out ``Post'' for the full 60-step budget until the trajectory is truncated, without recognizing the disabled state. The trained model, after two failed ``Post'' clicks, attends to the required-field asterisk on the \texttt{Add tags \& flair} entry, opens the flair selector, picks a flair (\texttt{Tech}), applies it, and submits in 22 steps (Figure~\ref{fig:reddit-case}). At step~15, the verbatim think trace of the trained model---translated from Chinese---explicitly verbalizes the bridging inference (``the button is still clickable but its color is gray, which may indicate the system has not detected all required fields\dots\ the `Add tags \& flair' entry has an asterisk indicating it is a required field''; full trace in Listing~\ref{lst:reddit-trained-think}).

The recovery occurs on a real-device gate condition not encountered in the training rollouts, illustrating Sim-to-Real OOD generalization at the level of individual interaction primitives. The structural barriers discussed in §\ref{sec:intro} make real-device online RL difficult to run at scale; simulator-based online RL on \sys{} offers a practical alternative. This case provides one instance in which the learned behavior transfers to a community-specific posting constraint.

\begin{lstlisting}[caption={Verbatim think trace on \texttt{Reddit\_CreatePostToCommunity}, translated from Chinese; ellipses indicate omitted boilerplate. The reasoning of the trained model at step~15 attends to the visual cue (grayed button, asterisked label) and re-plans; the base model emits byte-for-byte identical reasoning at every step from~10 onward.},label={lst:reddit-trained-think},basicstyle=\ttfamily\scriptsize,breaklines=true]
=== TRAINED MODEL (single real-device trial; 22 steps total) ===

[Step 13]  (first click on the grayed-out "Post" button)
  "Title and body are correctly filled. The 'Post' button in the
   top-right is in clickable state. Next: click 'Post' to publish."
  -> CLICK [880, 78]

[Step 14]  (second click on the grayed button; reasoning essentially
            identical to step 13)
  -> CLICK [880, 78]

[Step 15]  *** insight ***
  "Title and body are filled, all required information is ready.
   The 'Post' button is still clickable, BUT ITS COLOR IS GRAY,
   which may indicate the system has not detected all required
   fields (such as tags or labels) or other preconditions.
   ... The 'Add tags & flair' button on the current screen has an
   ASTERISK indicating it is a REQUIRED FIELD.
   Therefore, next: click 'Add tags & flair' to fill in the
   missing field, ensuring the post meets the publication
   requirements."
  -> CLICK [202, 255]      (the flair pill)

[Step 16]  (entered the tag selector)
  "Currently in the 'Add Tag' interface. 'None' is selected and the
   system indicates 'required'. ... 'Tech' is closer to the 'review'
   context, so select 'Tech'."
  -> CLICK [78, 668]       ('Tech' option)

[Step 17]  (after picking 'Tech')
  "'Tech' tag successfully selected, satisfying the required-field
   condition. The 'Apply' button is clickable. Click 'Apply'."
  -> CLICK [909, 128]

[Step 18]  (back at post page; 'Post' button now blue)
  "Title, body, and tag are all correctly filled. The 'Post' button
   is now blue, indicating it can be clicked. Click 'Post' to
   finally publish."
  -> CLICK [880, 78]       (succeeds)


=== BASE MODEL (single real-device trial; 60 steps = max budget) ===

[Step 10]
  "Title and body are filled, community is selected. Next: click
   'Post' to publish. The button is in the top-right."
  -> CLICK [880, 78]

[Steps 11..60]  (50 further responses)
  "Title and body are filled, community is selected. Next: click
   'Post' to publish. The button is in the top-right."
  -> CLICK [880, 78]
  (byte-for-byte identical reasoning across all 51 click steps; no
   recognition of the disabled state or the required-field asterisk)
\end{lstlisting}

\paragraph{The AnswerSheet protocol as a conservative gate.} Query/hybrid tasks in \sys{} end with the AnswerSheet protocol. The AnswerSheet \textbf{neither leaks the answer nor reduces the task to multiple choice}; it only adds an additional barrier of format and submission. On the test set, we observe a consistent pattern: on tasks containing an AnswerSheet, the simulation and real-device performance of the trained model are closely aligned (over all 19 AnswerSheet tasks: sim 71.1\% vs real 73.7\%; in the uplift subset of 12: sim 79.2\% vs real 83.3\%). For the base model, however, real-device free-text evaluation scores higher than simulation-side AnswerSheet evaluation (uplift subset: sim 2.1\% vs real 25.0\%). This pattern suggests that AnswerSheet imposes an additional submission-format barrier on weak models, and that \textbf{the AnswerSheet pass rate is a conservative lower bound of the free-text success rate}, not an inflated version of it. A strict ablation (toggling the AnswerSheet on the same rollout) is left to future work.

\section{Broader Uses of \sys{}}\label{app:broader-uses}

Beyond benchmarking, \sys{} enables several research directions that require controllable mobile environments.

\paragraph{Custom mobile environments and benchmarks.} Because apps, world data, task templates, and judges are modular, \sys{} can be extended beyond \bench{} to build targeted mobile benchmarks. Researchers can instantiate domain-specific environments, such as mobile finance, travel planning, social-media safety, or digital-literacy training, while retaining the same reset, snapshot, and state-based judging interface.

\paragraph{Controlled robustness and safety evaluation.} Programmable state and event injection make it possible to evaluate agents under systematic variations rather than incidental real-device conditions. The same task can be run under different balances, permissions, network states, incoming messages, popups, or phishing-like content. This supports controlled studies of robustness, prompt-injection susceptibility, caution gating, side effects, and recovery behavior.

\paragraph{Low-cost online RL research for GUI agents.} Because \sys{} can fork identical initial states into many lightweight browser instances, it provides a practical testbed for studying online RL in GUI environments without large emulator clusters or real-device farms. Researchers can compare reward designs, state-diff penalties, rollout grouping strategies, and Sim-to-Real behavior under reproducible initial states and deterministic outcome signals.

\paragraph{Controlled training data synthesis.} Each interaction step yields a five-tuple $(s_t^\mathrm{vis}, s_t^\mathrm{json}, a_t, s_{t+1}^\mathrm{vis}, s_{t+1}^\mathrm{json})$ of paired visual and structured state transitions. Because the environment is fully controllable, this data can be generated with intentional state coverage rather than incidental device logs, supporting training of mobile UI world models, state predictors, reward models, or trajectory verifiers.

\section{Detailed Footnotes for the Resource-Efficiency Comparison}\label{app:perf}

Detailed footnotes complementing the efficiency rows in Table~\ref{tab:comparison}:

\begin{itemize}
  \item The $\sim$50\,MB core disk footprint of \sys{} contains the framework code and the code of the 28 apps (JavaScript bundle, component code, state/navigation definitions, CSS, IME dictionary, and icon fonts). App content data (images, virtual-filesystem presets, in-app corpora, etc.) scales linearly with the number of apps and content richness as an optional layer; the current full deployment is around 1.5\,GB and can be slimmed or replaced as needed. Android-emulator environments are dominated by the Android system image, which is largely independent of benchmark task count or content data.

  \item AndroidWorld README states that the emulator guest memory is 2\,GB; in our Docker measurements, host occupation steadily sits at $\sim$4.5\,GB (containing emulator + FastAPI server + Android 13 system image, without \texttt{/dev/kvm}).

  \item The Docker image of AndroidWorld totals 20.2\,GB, of which the Android 13 system image accounts for 9.5\,GB. Multiple emulator instances can share the same image, but each instance still has a $\sim$1\,GB \texttt{userdata.img}.

  \item MobileWorld is built on top of the AndroidWorld emulator stack; we therefore report its memory and disk figures as lower bounds ($\geq$4.5\,GB and $\geq$20\,GB) inherited from the AndroidWorld baseline rather than direct measurements. The actual figures cannot fall below this baseline but were not separately benchmarked. AndroidLab's numbers ($\sim$6\,GB, $\sim$9\,GB) are taken from its repository's stated emulator configuration; we did not independently measure them.

  \item The emulator boot of AndroidWorld has a measured median of 78\,s without \texttt{/dev/kvm}; with KVM enabled, it is usually faster. After boot, the FastAPI server still has to perform automatic app setup (Chrome, Contacts, etc.), and the time-to-fully-ready can reach the minute level in our test environment.

  \item The \texttt{/reset} endpoint of AndroidWorld does not reboot the emulator or wipe app data; it only performs \texttt{press\_home} + clear interaction\_cache. Task-level state restoration is achieved via \texttt{app\_snapshot} (file-level copy of \texttt{/data/data/<package>}), which is constrained by what the OS surfaces at the file-system layer. App-internal in-memory state and account/backend state are not captured. \sys{} restores state via direct JSON setState injection into the same in-memory stores that the apps read from, so the restoration scope matches the state the agent can affect.

  \item The AndroidWorld repository does not provide an in-process multi-session runner; running multiple sessions concurrently requires launching an independent Docker container per session (with its own emulator, app data, snapshot, and host port), so resource overhead scales linearly. \sys{} can directly manage multiple browser contexts / pages within a single process via \texttt{EnvPool}, with no duplicated OS or system-image overhead.
\end{itemize}

\section{Cost Table if Switching to a VLM Judge}\label{app:cost}

To anchor the abstract cost argument to concrete scenarios, we use \textbf{one full evaluation run on the 256-task \sys{} test set} as the unit and compute the VLM-judge API cost in two typical scenarios. The estimate is based on a sampled VLM audit of 546 screenshots, where each trajectory consumes on average $\sim$29.8K input tokens including screenshots and $\sim$924 output tokens.

\begin{table}[!htbp]
  \centering\small
  \setlength{\tabcolsep}{3pt}
  \resizebox{\columnwidth}{!}{%
  \begin{tabular}{lcc}
    \toprule
    \textbf{Scenario} & \textbf{Qwen3.6-Plus$^\dagger$} & \textbf{GPT-5.4$^\ddagger$} \\
    \midrule
    One eval over 1$\times$256 tasks & $\sim$\$2.6 (\textyen 18) & $\sim$\$23 (\textyen 158) \\
    GRPO 1 step (96 traj)            & $\sim$\$1 (\textyen 7)    & $\sim$\$8.5 (\textyen 59) \\
    \midrule
    \multicolumn{3}{l}{\textbf{Code-level judging (this paper)}: \$0 (any scale)} \\
    \bottomrule
  \end{tabular}%
  }
  \caption{Per-run cost comparison if a VLM judge were used}
  \label{tab:cost-actual}
\end{table}

$^\dagger$Aliyun Bailian pricing \textyen 2/M input + \textyen 12/M output (within 256K). $^\ddagger$OpenAI API pricing for GPT-5.4: \$2.50/M input + \$15/M output, converted at a 7$\times$ exchange rate. The GPT-5.4 path is roughly 8.75$\times$ as expensive as the Qwen path.

\begin{table}[!htbp]
  \centering\small
  \setlength{\tabcolsep}{3pt}
  \resizebox{\columnwidth}{!}{%
  \begin{tabular}{lccc}
    \toprule
    \textbf{Scale} & \textbf{Qwen3.6-Plus} & \textbf{GPT-5.4} & \textbf{Code-level} \\
    \midrule
    100 step (9.6K traj)      & $\sim$\$100   & $\sim$\$850   & \$0 \\
    1K step (96K traj)        & $\sim$\$1,000 & $\sim$\$8.5K  & \$0 \\
    10K step (960K traj)$^*$  & $\sim$\$10K   & $\sim$\$85K   & \$0 \\
    \bottomrule
  \end{tabular}%
  }
  \caption{Cumulative cost at large-scale RL training if a VLM judge were used}
  \label{tab:cost-scale}
\end{table}

$^*$Comparable to the ``millions of interactive rollouts'' training scale publicly reported by UI-TARS-2~\cite{UITARS2}.

We emphasize that the above is only the \textbf{VLM-judge API cost}: a complete real-environment RL training run also incurs the cost of cloud devices / emulator rentals. GUI-Genesis~\cite{GUI-GENESIS} reports that, in their WeChat mini-program experiments, the real-environment + VLM-reward configuration costs as much as \$240 per step (\$0.17/min cloud-device rental + \$0.005/trajectory VLM verification), and a single epoch (1K env $\times$ 12 rollout = 12K trajectories) costs approximately \textbf{\$28,000}; infrastructure costs clearly dominate in that setting. Because \sys{} runs as a browser environment on local machines, it avoids this category of infrastructure cost in our setup.

\end{document}